\documentclass{article}
\usepackage{mtfuse} % I've modified KR'99's proceedings.sty to
\usepackage{latexsym}
\usepackage{graphics}
\newcommand{\email}[1]{{\tt #1}}
\newcommand{\url}[1]{{\tt #1}}
\newcommand{\keywords}[1]{{\bf Keywords:} #1}

\newcommand{\comment}[1]{}

\newcommand{\mi}[1]{\mathit{#1}}

\newcommand{\mca}{{\mathcal{A}}}
\newcommand{\mcb}{{\mathcal{B}}}
\newcommand{\mcl}{{\mathcal{L}}}

\newcommand{\mcr}{{\mathcal{R}}}
\newcommand{\mcq}{{\mathcal{Q}}}
\newcommand{\mcsq}{{\mathcal{Q_<}}}
\newcommand{\mcs}{{\mathcal{S}}}
\newcommand{\mct}{{\mathcal{T}}}
\newcommand{\mcst}{{\mathcal{T_<}}}
\newcommand{\mcw}{{\mathcal{W}}}

\newcommand{\C}{C}
\newcommand{\agnost}{\sim}
\newcommand{\sagnost}{\approx}
\newcommand{\conflict}{\infty}
\newcommand{\notconflict}{\mbox{\makebox[0pt][l]{$\not$}}\conflict}
\newcommand{\sconflict}{\bowtie}

\newcommand{\fuse}{\bigcirc\hspace*{-0.8em}{\small\vee}\ }
\newcommand{\implies}{\Rightarrow}
\newcommand{\sqeq}{\equiv}

\newcommand{\qed}{\rule{2mm}{2mm}}

\newtheorem{prop}{Proposition}
\newtheorem{defn}{Definition}
\newtheorem{exmp}{Example}

\newtheorem{lem}{Lemma}[prop]

\newenvironment{pf}{\trivlist \item[\hskip \labelsep{\bf
Proof: }]}{ \endtrivlist}

\begin{document}

\title{
	Representing and Aggregating Conflicting Beliefs}
\author{
	{\bf Pedrito Maynard-Reid II} \\
	Department of Computer Science \\
	Stanford University \\
	Stanford, CA 94305, USA \\
	\email{pedmayn@cs.stanford.edu}
\And
	{\bf Daniel Lehmann} \\
	School of Computer Science and Engineering \\
	Hebrew University \\
	Jerusalem 91904, Israel \\
	\email{lehmann@cs.huji.ac.il}} 
%\date{Started: 1999 March 25, Thursday}
%\date{}

\pubnote{Shorter ver.\ in {\em Proc.\ 7th Intl.\ Conf.\ on Principles
of Knowledge Representation and Reasoning (KR 2000)}.}

\maketitle

\begin{abstract}
We consider the two-fold problem of representing collective beliefs
and aggregating these beliefs. We propose modular, transitive
relations for collective beliefs. They allow us to represent
conflicting opinions and they have a clear semantics.  We compare them
with the quasi-transitive relations often used in Social
Choice. Then, we describe a way to construct the belief state of an
agent informed by a set of sources of varying degrees of
reliability. This construction circumvents Arrow's Impossibility
Theorem in a satisfactory manner. Finally, we give a simple
set-theory-based operator for combining the information of multiple
agents. We show that this operator satisfies the desirable invariants
of idempotence, commutativity, and associativity, and, thus, is
well-behaved when iterated, and we describe a computationally
effective way of computing the resulting belief state.
\end{abstract}


\section{Introduction}\label{sec-intro}
%	 ____________

We are interested in the multi-agent setting where agents are informed
by sources of varying levels of reliability, and where agents can
iteratively combine their belief states. This setting introduces three
problems: (1) Finding an appropriate representation for collective
beliefs; (2) Constructing an agent's belief state by aggregating the
information from informant sources, accounting for the relative
reliability of these sources; and, (3) Combining the information of
multiple agents in a manner that is well-behaved under iteration.

The Social Choice community has dealt extensively with the first
problem (although in the context of representing collective
preferences rather than beliefs) (cf.~\cite{Sen86}). The classical
approach has been to use quasi-transitive relations (of which total
pre-orders are a special subclass) over the set of possible
worlds. However, these relations do not distinguish between group
indifference and group conflict, and this distinction can be
crucial. Consider, for example, a situation in which all members of a
group are indifferent between movie $a$ and movie $b$.  If some
passerby expresses a preference for $a$, the group may very well
choose to adopt this opinion for the group and borrow $a$.  However,
if the group was already divided over the relative merits of $a$ and
$b$, we would be wise to hesitate before choosing one over the other
just because a new supporter of $a$ appears on the scene. We propose a
representation in which the distinction is explicit. We also argue
that our representation solves some of the unpleasant semantical
problems suffered by the earlier approach.

The second problem addresses how an agent should actually go about
combining the information received from a set of sources to create a
belief state. Such a mechanism should favor the opinions held by more
reliable sources, yet allow less reliable sources to voice opinions
when higher ranked sources have no opinion.  True, under some
circumstances it would not be advisable for an opinion from a less
reliable source to override the agnosticism of a more reliable source,
but often it is better to accept these opinions as default assumptions
until better information is available. \cite{MS00} provides a solution
to this problem when belief states are represented as total pre-orders,
but runs into Arrow's Impossibility Theorem~\cite{Arrow63} when there
are sources of equal reliability. As we shall see, the generalized
representation allows us to circumvent this limitation.

To motivate the third problem, consider the following dynamic
scenario: A robot controlling a ship in space receives from a number
of communication centers on Earth information about the status of its
environment and tasks. Each center receives information from a group
of sources of varying credibility or accuracy (e.g., nearby satellites
and experts) and aggregates it. Timeliness of decision-making in space
is often crucial, so we do not want the robot to have to wait while
each center sends its information to some central location for it to
be first combined before being forwarded to the robot. Instead, each
center sends its aggregated information directly to the robot. Not
only does this scheme reduce dead time, it also allows for ``anytime''
behavior on the robot's part: the robot incorporates new information
as it arrives and makes the best decisions it can with whatever
information it has at any given point. This distributed approach is
also more robust since the degradation in performance is much more
graceful should information from individual centers get lost or
delayed.

In such a scenario, the robot needs a mechanism for combining or {\em
fusing} the belief states of multiple agents potentially arriving at
different times. Moreover, the belief state output by the mechanism
should be invariant with respect to the order of agent arrivals. We
will describe such a mechanism.

The paper is organized as follows: After some preliminary definitions
and a discussion of the approach to aggregation taken in classical
Social Choice, we introduce modular, transitive relations for
representing generalized belief states. We then describe how to
construct the belief state of an agent given the belief states of its
informant sources when these sources are totally pre-ordered. Finally,
we describe a simple set-theory-based operator for fusing agent belief
states that satisfies the desirable invariants of idempotence,
commutativity, and associativity, and we describe a computationally
effective way of computing this belief state.

\section{Preliminaries}
%	 _____________

We begin by defining various well-known properties of binary
relations\footnote{We only use binary relations in this paper, so we
will refer to them simply as relations.}; they will be useful to us
throughout the paper.
\begin{defn}
	Suppose $\leq$ is a relation over a finite set $\Omega$, i.e.,
	\mbox{$\leq \subseteq \Omega \times \Omega$}. We shall use
	\mbox{$x \leq y$} to denote \mbox{$(x, y) \in \leq$} and
	\mbox{$x \not\leq y$} to denote \mbox{$(x, y) \not\in
	\leq$}. The relation $\leq$ is:
	\begin{enumerate}

	\item {\em reflexive} iff \mbox{$x \leq x$} for \mbox{$x \in
	\Omega$}. It is {\em irreflexive} iff \mbox{$x \not\leq x$}
	for \mbox{$x \in \Omega$}.

	\item {\em symmetric} iff \mbox{$x \leq y \implies y \leq x$}
	for \mbox{$x, y \in \Omega$}. It is {\em asymmetric} iff
	\mbox{$x \leq y \implies y \not\leq x$} for \mbox{$x, y \in
	\Omega$}. It is {\em anti-symmetric} iff \mbox{$x \leq y
	\wedge y \leq x \implies x = y$} for \mbox{$x, y \in \Omega$}.

	\item the {\em strict version} of a relation $\leq'$ over
	$\Omega$ iff \mbox{$x \leq y \Leftrightarrow x \leq' y \wedge
	y \not\leq' x$} for \mbox{$x, y \in \Omega$}.

	\item {\em total} iff \mbox{$x \leq y \vee y \leq x$} for
	\mbox{$x, y \in \Omega$}.

	\item {\em modular} iff \mbox{$x \leq y \implies x \leq z \vee
	z \leq y$} for \mbox{$x, y, z \in \Omega$}.

	\item {\em transitive} iff \mbox{$x \leq y \wedge y \leq z
	\implies x \leq z$} for \mbox{$x, y, z \in \Omega$}.

	\item {\em quasi-transitive} iff its strict version is
	transitive.

	\item the {\em transitive closure} of a relation $\leq'$ over
	$\Omega$ iff \mbox{$x \leq y$} \mbox{$\Leftrightarrow$}
	\mbox{$\exists w_0, \ldots, w_n \in \Omega.$} \mbox{$x = w_0
	\leq' \cdots \leq' w_n = y$} for some integer $n$, for
	\mbox{$x, y \in \Omega$}.

	\item {\em acyclic} iff \mbox{$\forall w_0, \ldots, w_n \in
	\Omega.\ w_0 < \cdots < w_n$} implies \mbox{$w_n \not< w_0$}
	for all integers $n$, where $<$ is the strict version of
	$\leq$.

	\item a {\em total pre-order} iff it is total and
	transitive. It is a {\em total order} iff it is also
	anti-symmetric.

	\item an {\em equivalence relation} iff it is reflexive,
	symmetric, and transitive.

	\end{enumerate}

\end{defn}

\begin{prop}\label{prop-Props}\hspace{0ex}\\[-0.5cm]
	\begin{enumerate}

	\item The transitive closure of a modular relation is
	modular.

	\item Every transitive relation is quasi-transitive.

	\item \cite{Sen86} Every quasi-transitive relation is
	acyclic.

	\end{enumerate}
\end{prop}

Given a relation over a set of alternatives and a subset of these
alternatives, we often want to pick the subset's ``best'' elements
with respect to the relation. We define this set of ``best'' elements
to be the subset's {\em choice set}:
\begin{defn}
	If $\leq$ is a relation over a finite set $\Omega$, $<$ is its
	strict version, and \mbox{$X \subseteq \Omega$}, then the {\em
	choice set of $X$ with respect to $\leq$} is
	\begin{displaymath}
		\C(X, \leq) = \{x \in X : \not\exists x' \in X.\ x' <
			x\}.
	\end{displaymath}
\end{defn}
A {\em choice function} is one which assigns to every subset $X$ a
non-empty subset of $X$:
\begin{defn}
	A {\em choice function} over a finite set $\Omega$ is a
	function \mbox{$f : {2^\Omega}\setminus\emptyset \to
	{2^\Omega}\setminus\emptyset$} such that \mbox{$f(X) \subseteq
	X$} for every \mbox{$X \subseteq \Omega$}.
\end{defn}
Now, every acyclic relation defines a choice function, one which
assigns to each subset its choice set:
\begin{prop}\label{prop-Choice}
	\cite{Sen86} Given a relation $\leq$ over a finite set
	$\Omega$, the choice set operation $\C$ defines a choice
	function iff $\leq$ is acyclic.\footnote{Sen's uses a slightly
	stronger definition of choice sets, but the theorem still
	holds in our more general case.}
\end{prop}
If a relation is not acyclic, elements involved in a cycle are said to
be in a {\em conflict} because we cannot order them:
\begin{defn}\label{def-conflict}
	Given a relation $<$ over a finite set $\Omega$, $x$ and $y$
	are in a {\em conflict wrt $<$} iff there exist \mbox{$w_0,
	\ldots, w_n, z_0, \ldots, z_m \in \Omega$} such that \mbox{$x
	= w_0 < \cdots < w_n = y = z_0 < \cdots < z_m = x$}, where
	\mbox{$x, y \in \Omega$}.
\end{defn}

\section{Aggregation in Social Choice}
%	 ____________________________

We are interested in belief aggregation, but the community
historically most interested in aggregation has been that of Social
Choice theory. The aggregation is over preferences rather than
beliefs, so the discussion in this subsection will focus on
representing preferences; however, as we shall see, the results are
equally relevant to representing beliefs. In the Social Choice
community, the standard representation of an agent's preferences is a
total pre-order. Each total pre-order $\preceq_i$ is interpreted as
describing the weak preferences of an individual $i$, so that \mbox{$x
\preceq_i y$} means $i$ considers alternative $x$ to be at least as
preferable as alternative $y$.\footnote{The direction of the relation
symbol is unintuitive, but standard practice in the belief revision
community.}  If \mbox{$x \preceq_i y$} and \mbox{$y \preceq_i x$},
then $i$ is indifferent between $x$ and $y$.

Unfortunately, Arrow's Impossibility Theorem~\cite{Arrow63} showed
that no aggregation operator over total pre-orders exists satisfying
the following small set of desirable properties:
\begin{defn}
Let $f$ be an aggregation operator over the preferences $\preceq_1$,
\ldots, $\preceq_n$ of $n$ individuals, respectively, over a finite
set of alternatives $\Omega$, and let \mbox{$\preceq\ = f(\preceq_1,
\ldots, \preceq_n)$}.
	\begin{itemize}

	\item {\em Restricted Range}: The range of $f$ is the set of
	total pre-orders over $\Omega$.

	\item {\em Unrestricted Domain}: The domain of $f$ is the set
	of $n$-tuples of total pre-orders over $\Omega$.

	\item {\em Pareto Principle}: If \mbox{$x \prec_i y$} for all
	$i$, then \mbox{$x \prec y$}.

	\item {\em Independence of Irrelevant Alternatives (IIA)}:
	Suppose \mbox{$\preceq'\ = f(\preceq_1', \ldots,
	\preceq_n')$}. If, for \mbox{$x, y \in \Omega$}, \mbox{$x
	\preceq_i y$} iff \mbox{$x \preceq_i' y$} for all $i$, then
	\mbox{$x \preceq y$} iff \mbox{$x \preceq' y$}.

	\item {\em Non-Dictatorship}: There is no individual $i$ such
	that, for every tuple in the domain of $f$ and every \mbox{$x,
	y \in \Omega$}, \mbox{$x \prec_i y$} implies \mbox{$x \prec
	y$}.

	\end{itemize}
\end{defn}

\begin{prop}\label{prop-Arrow}
	\cite{Arrow63} There is no aggregation operator that satisfies
	restricted range, unrestricted domain, (weak) Pareto
	principle, independendence of irrelevant alternatives, and
	nondictatorship.
\end{prop}
This impossibility theorem led researchers to look for weakenings to
Arrow's framework that would circumvent the result. One was to weaken
the restricted range condition, requiring that the result of an
aggregation only satisfy totality and quasi-transitivity rather than
the full transitivity of a total pre-order. This weakening was
sufficient to guarantee the existence of an aggregation function
satisfying the other conditions, while still producing relations that
defined choice functions~\cite{Sen86}. However, this solution was not
without its own problems.

First, total, quasi-transitive relations have unsatisfactory
semantics.  If $\preceq$ is total and quasi-transitive but not a total
pre-order, its indifference relation is not transitive:
\begin{prop}
	Let $\preceq$ be a relation over a finite set $\Omega$ and let
	$\agnost$ be its symmetric restriction (i.e., \mbox{$x \agnost
	y$} iff \mbox{$x \preceq y$} and \mbox{$y \preceq x$}). If
	$\preceq$ is total and quasi-transitive but not transitive,
	then $\agnost$ is not transitive.
\end{prop}
There has been much discussion as to whether or not indifference
should be transitive; in many cases one feels indifference should be
transitive. If Deb enjoys plums and mangoes equally and also enjoys
mangoes and peaches equally, we would conclude that she also enjoys
plums and peaches equally. It seems that total quasi-transitive
relations that are not total pre-orders cannot be understood easily as
preference or indifference.

Since the existence of a choice function is generally sufficient for
classical Social Choice problems, this issue was at least
ignorable. However, in iterated aggregation, the result of the
aggregation must not only be usable for making decisions, but must be
interpretable as a new preference relation that may be involved in
later aggregations; consequently, it must maintain clean semantics.

Secondly, the totality assumption is excessively restrictive for
representing aggregate preferences. In general, a binary relation
$\preceq$ can express four possible relationships between a pair of
alternatives $a$ and $b$: \mbox{$a \preceq b$} and \mbox{$b
\not\preceq a$}, \mbox{$b \preceq a$} and \mbox{$a \not\preceq b$},
\mbox{$a \preceq b$} and \mbox{$b \preceq a$}, and \mbox{$a
\not\preceq b$} and \mbox{$b \not\preceq a$}. Totality reduces this
set to the first three which, under the interpretation of relations as
representing weak preference, correspond to the two strict orderings
of $a$ and $b$, and indifference. However, consider the situation
where a couple is trying to choose between an Italian and an Indian
restaurant, but one strictly prefers Italian food to Indian food,
whereas the second strictly prefers Indian to Italian. The couple's
opinions are in conflict, a situation that does not fit into any of
the three remaining categories. Thus, the totality assumption is
essentially an assumption that conflicts do not exist. This, one may
argue, is appropriate if we want to represent preferences of one agent
(but see~\cite{KT79} for persuasive arguments that individuals {\em
are} often ambivalent).  However, the assumption is inappropriate if
we want to represent aggregate preferences since individuals will
almost certainly have differences of opinion.

\section{Generalized Belief States}
%	 _________________________

Let us turn to the domain of belief aggregation. A total pre-order over
the set of possible worlds is a fairly well-accepted representation
for a belief state in the belief revision
community~\cite{Grove88,KM91,LM92,GM94}. Instead of preference,
relations represent relative likelihood, instead of indifference,
equal likelihood. For the remainder of the paper, assume we are given
some language $\mcl$ with a satisfaction relation $\models$ for
$\mcl$. Let $\mcw$ be a finite, non-empty set of possible worlds
(interpretations) over $\mcl$. Suppose $\preceq$ is a total pre-order
on $\mcw$. The belief revision literature maintains that the
conditional belief ``if $p$ then $q$'' (where $p$ and $q$ are
sentences in $\mcl$) holds if all the worlds in the choice set of
those satisfying $p$ also satisfy $q$; we write \mbox{$Bel(p ?
q)$}. The individual's unconditional beliefs are all those where $p$
is the sentence \mbox{$true$}. If neither the belief \mbox{$p ?  q$}
nor its negation hold in the belief state, it is said to be {\em
agnostic} with respect to \mbox{$p ? q$}, written \mbox{$Agn(p ? q)$}.

It should come as no surprise that belief aggregation is formally
similar to preference aggregation and, as a result, is also
susceptible to the problems described in the previous section. We
propose a solution to these problems which generalizes the total
pre-order representation so as to capture information about conflicts.

\subsection{Modular, transitive states}
%	    --------------------------

We take strict likelihood as primitive. Since strict likelihood is not
necessarily total, it is possible to represent agnosticism and
conflicting opinions in the same structure. This choice deviates from
that of most authors, but are similar to those of
Kreps~\cite[p.~19]{Kreps90} who is interested in representing both
indifference and incomparability. Unlike Kreps, rather than use an
asymmetric relation to represent strict likelihood (e.g., the strict
version of a weak likelihood relation), we impose the less restrictive
condition of modularity.

We formally define {\em generalized belief states}:
\begin{defn}
	A {\em generalized belief state $\prec$} is a modular,
	transitive relation over $\mcw$. The set of possible
	generalized belief states over $\mcw$ is denoted $\mcb$.
\end{defn}
We interpret \mbox{$a \prec b$} to mean ``there is reason to consider
$a$ as strictly more likely than $b$.'' We represent equal likelihood,
which we also refer to as ``agnosticism,'' with the relationship
$\agnost$ defined such that \mbox{$x \agnost y$} if and only if
\mbox{$x \not\prec y$} and \mbox{$y \not\prec x$}. We define the
conflict relation corresponding to $\prec$, denoted $\conflict$, so
that \mbox{$x \conflict y$} iff \mbox{$x \prec y$} and \mbox{$y \prec
x$}. It describes situations where there are reasons to consider
either of a pair of worlds as strictly more likely than the other. In
fact, one can easily check that $\conflict$ precisely represents
conflicts in a belief state in the sense of
Definition~\ref{def-conflict}.

For convenience, we will refer to generalized belief states simply as
belief states for the remainder of the paper except when to do so
would cause confusion.

\subsection{Discussion}
%	    ----------

Let us consider why our choice of representation is justified. First,
we agree with the Social Choice community that strict likelihood
should be transitive.

As we discussed in the previous section, there is often no compelling
reason why agnosticism/indifference should not be transitive; we also
adopt this view. However, transitivity of strict likelihood by itself
does not guarantee transitivity of agnosticism. A simple example is
the following: \mbox{$\prec = \{(a,c)\}$}, so that \mbox{$\agnost =
\{(a,b), (b,c)\}$}. However, if we buy that strict likelihood should
be transitive, then agnosticism is transitive identically when strict
likelihood is also modular:
\begin{prop}
	Suppose a relation $\prec$ is transitive and $\agnost$ is the
	corresponding agnosticism relation. Then $\agnost$ is
	transitive iff $\prec$ is modular.
\end{prop}
In summary, transitivity and modularity are necessary if strict
likelihood and agnosticism are both required to be transitive.

We should point out that conflicts are also transitive in our
framework. At first glance, this may appear undesirable: it is
entirely possible for a group to disagree on the relative likelihood
of worlds $a$ and $b$, and $b$ and $c$, yet agree that $a$ is more
likely than $c$. However, we note that this transitivity follows from
the cycle-based definition of conflicts
(Definition~\ref{def-conflict}), not from our belief state
representation. It highlights the fact that we are not only concerned
with conflicts that arise from simple disagreements over pairs of
alternatives, but those that can be inferred from a series of
inconsistent opinions as well.

Now, to argue that modular, transitive relations are sufficient to
capture relative likelihood, agnosticism, and conflicts among a group
of information sources, we first point out that adding irreflexivity
would give us the class of relations that are strict versions of total
pre-orders, i.e., conflict-free. Let $\mct$ be the set of total
pre-orders over $\mcw$, $\mcst$, the set of their strict versions.
\begin{prop}\label{prop-TPisoMTI}
	The set of irreflexive relations in $\mcb$ is isomorphic to
	$\mct$ and, in fact, equals $\mcst$.
\end{prop}

\begin{figure}[htb]
\centerline{\resizebox{!}{1in}{\includegraphics{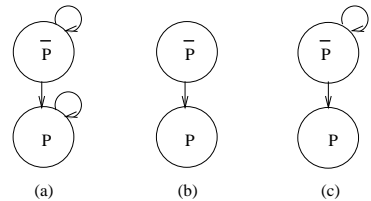}}}
%\centerline{\epsfig{file=gbs.eps,height=1in}}
\caption{Three examples of generalized belief states: (a) a total
pre-order, (b) the strict version of a total pre-order, (c)
neither. (Each circle represents all the worlds in $\mcw$ which
satisfy the sentence inside. An arc between circles indicates that $w
\prec w'$ for every $w$ in the head circle and $w'$ in the tail
circle; no arc indicates that $w \not\prec w'$ for each of these
pairs. In particular, the set of worlds represented by a circle is
fully connected if there is an arc from the circle to itself, fully
disconnected otherwise.)}\label{fig-gbs}
\end{figure}

Secondly, the following representation theorem shows that each belief
state partitions the possible worlds into sets of worlds either all
equally likely or all potentially involved in a conflict, and totally
orders these sets; worlds in distinct sets have the same relation to
each other as do the sets.
\begin{prop}\label{prop-rep}
	\mbox{$\prec \in \mcb$} iff there is a partition
	\mbox{$\mathbf{W} = \langle W_0, \ldots, W_n \rangle$} of
	$\mcw$ such that:
	\begin{enumerate}

	\item For every \mbox{$x \in W_i$} and \mbox{$y \in W_j$},
	\mbox{$i \neq j$} implies \mbox{$i < j$} iff \mbox{$x \prec
	y$}.

	\item Every $W_i$ is either fully connected (\mbox{$w \prec
	w'$} for all \mbox{$w, w' \in W_i$}) or fully disconnected
	(\mbox{$w \not\prec w'$} for all \mbox{$w, w' \in W_i$}).

	\end{enumerate}
\end{prop}
Figure~\ref{fig-gbs} shows three examples of belief states: one which is a
total pre-order, one which is the strict version of a total pre-order,
and one which is neither.

Thus, generalized belief states are not a big change from the strict
versions of total pre-orders. They merely generalize these by weakening
the assumption that sets of worlds not strictly ordered are equally
likely, allowing for the possibility of conflicts. Now we can
distinguish between agnostic and conflicting conditional beliefs. A
belief state $\prec$ is agnostic about conditional belief \mbox{$p ?
q$} (i.e., \mbox{$Agn(p ?  q)$}) if the choice set of worlds
satisfying $p$ contains both worlds which satisfy $q$ and $\neg q$ and
is fully disconnected. It is in conflict about this belief, written
\mbox{$Con(p ? q)$}, if the choice set is fully connected.

Finally, we compare the representational power of our definitions to
those discussed in the previous section. First, $\mcb$ subsumes the
class of total pre-orders:
\begin{prop}\label{prop-TPsubMT}
	\mbox{$\mct \subset \mcb$} and is the set of reflexive
	relations in $\mcb$.
\end{prop}
Secondly, $\mcb$ neither subsumes nor is subsumed by the set of total,
quasi-transitive relations, and the intersection of the two classes is
$\mct$. Let $\mcq$ be the set of total, quasi-transitive relations
over $\mcw$, and $\mcsq$, the set of their strict versions.
\begin{prop}\label{prop-MTandTQ}\hspace{0ex}\\[-0.5cm]
	\begin{enumerate}

	\item \mbox{$\mcq \cap \mcb = \mct$}.

	\item \mbox{$\mcb \not\subseteq \mcq$}.

	\item \mbox{$\mcq \not\subseteq \mcb$} if $\mcw$ has at least
	three elements.

	\item \mbox{$\mcq \subset \mcb$} if $\mcw$ has one or two
	elements.

	\end{enumerate}
\end{prop}
Because modular, transitive relations represent strict preferences, it
is probably fairer to compare them to the class of strict versions of
total, quasi-transitive relations. Again, neither class subsumes the
other, but this time the intersection is $\mcst$:
\begin{prop}\hspace{0ex}\\[-0.5cm]
	\begin{enumerate}

	\item \mbox{$\mcsq \cap \mcb = \mcst$}.

	\item \mbox{$\mcb \not\subseteq \mcsq$}.

	\item \mbox{$\mcsq \not\subseteq \mcb$} if $\mcw$ has at least
	three elements.

	\item \mbox{$\mcsq \subset \mcb$} if $\mcw$ has one or two
	elements.

	\end{enumerate}
\end{prop}

In the next section, we define a natural aggregation policy based on
this new representation that admits clear semantics and obeys
appropriately modified versions of Arrow's conditions.

\section{Single-agent belief state construction}\label{sec-AGR}
%	 ______________________________________

Suppose an agent is informed by a set of sources, each with its
individual belief state. Suppose further that the agent has ranked the
sources by level of credibility. We propose an operator for
constructing the agent's belief state $\prec$ by aggregating the
belief states of the sources in $S$ while accounting for the
credibility ranking of the sources.

\begin{exmp}\label{ex-robot}
	We will use a running example from our space robot domain to
	help provide intuition for our definitions. The robot sends to
	earth a stream of telemetry data gathered by the spacecraft,
	as long as it receives positive feedback that the data is
	being received. At some point it loses contact with the
	automatic feedback system, so it sends a request for
	information to an agent on earth to find out if the failure
	was caused by a failure of the feedback system or by an
	overload of the data retrieval system. In the former case, it
	would continue to send data, in the latter, desist. As it so
	happens, there has been no overload, but the computer running
	the feedback system has hung. The agent consults the following
	three experts, aggregates their beliefs, and sends the results
	back to the robot:
	\begin{enumerate}

	\item $s_p$, the computer programmer that developed the
	feedback program, believes nothing could ever go wrong with
	her code, so there must have been an overload
	problem. However, she admits that if her program had crashed,
	the problem could ripple through to cause an overload.

	\item $s_m$, the manager for the telemetry division,
	unfortunately has out-dated information that the feedback
	system is working. She was also told by the engineer who sold
	her the system that overloading could never happen. She has no
	idea what would happen if there was an overload or the
	feedback system crashed.

	\item $s_t$, the technician working on the feedback system,
	knows that the feedback system crashed, but doesn't know
	whether there was a data-overload. Not being familiar with the
	retrieval system, she is also unable to speculate whether the
	data retrieval system would have overloaded if the feedback
	system had not failed.

	\end{enumerate}

	Let $F$ and $D$ be propositional variables representing that
	the feedback and data retrieval systems, respectively, are
	okay. The belief states for the three sources are shown in
	Figure~\ref{fig-robot}.

	\begin{figure}[htb]
	\centerline{\resizebox{!}{2in}{\includegraphics{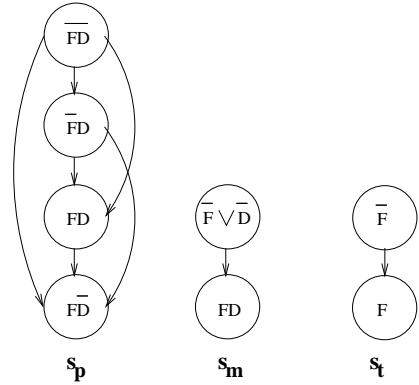}}}
	%\centerline{\epsfig{file=robot.eps,height=2in}}
	\caption{The belief states of $s_p$, $s_m$, and $s_t$ in
	Example~\ref{ex-robot}.}\label{fig-robot}
	\end{figure}
\end{exmp}

Let us begin the formal development by defining sources:
\begin{defn}\label{defn-sources}
	$\mcs$ is a finite set of {\em sources}. With each source
	\mbox{$s \in \mcs$} is associated a belief state \mbox{$<^s
	\in \mcb$}.
\end{defn}
We denote the agnosticism and conflict relations of a source $s$ by
$\sagnost^s$ and $\sconflict^s$, respectively.  It is possible to
assume that the belief state of a source is conflict free, i.e.,
acyclic. However, this is not necessary if we allow sources to suffer
from the human malady of ``being torn between possibilities.''

We assume that the agent's credibility ranking over the sources is a
total pre-order:
\begin{defn}
	$\mcr$ is a totally ordered finite set of {\em ranks}.
\end{defn}

\begin{defn}
	\mbox{$\mi{rank} : \mcs \to \mcr$} assigns to each source a
	rank.
\end{defn}

\begin{defn}
	$\sqsupseteq$ is the total pre-order over $\mcs$ induced by the
	ordering over $\mcr$. That is, \mbox{$s \sqsupseteq s'$} iff
	\mbox{$rank(s) \geq rank(s')$}; we say $s'$ is {\em as
	credible as} $s$.  $\sqsupseteq_S$ is the restriction of
	$\sqsupseteq$ to \mbox{$S \subseteq \mcs$}.
\end{defn}
We use $\sqsupset$ and $\sqeq$ to denote the asymmetric and symmetric
restrictions of $\sqsupseteq$, respectively.\footnote{Note that,
unlike the relations representing belief states, $\geq$ and
$\sqsupseteq$ are read in the intuitive way, that is, ``greater''
corresponds to ``better.''} The finiteness of $\mcs$ ($\mcr$) ensures
that a maximal source (rank) always exists, which is necessary for
some of our results. Weaker assumptions are possible, but at the price
of unnecessarily complicating the discussion.

We are ready to consider the source aggregation problem. In the
following, assume an agent is informed by a set of sources \mbox{$S
\subseteq \mcs$}. We look at two special cases---equal-ranked and
strictly-ranked source aggregation---before considering the general
case.

\subsection{Equal-ranked sources aggregation}
%	    --------------------------------

Suppose all the sources have the same rank so that $\sqsupseteq_S$ is
fully connected. Intuitively, we want take all offered opinions
seriously, so we take the union of the relations:
\begin{defn}
	If \mbox{$S \subseteq \mcs$}, then \mbox{$Un(S)$} is the
	relation \mbox{$\bigcup_{s \in S} <^s$}.
\end{defn}
By simply taking the union of the source belief states, we may lose
transitivity. However, we do not lose modularity:
\begin{prop}
	If \mbox{$S \subseteq \mcs$}, then \mbox{$Un(S)$} is modular
	but not necessarily transitive.
\end{prop}
Thus, we know from Proposition~\ref{prop-Props} that we need only take
the transitive closure of \mbox{$Un(S)$} to get a belief state:
\begin{defn}
	If \mbox{$S \subseteq \mcs$}, then \mbox{$AGRUn(S)$} is the
	relation \mbox{$Un(S)^+$}.
\end{defn}

\begin{prop}\label{prop-AGRUn}
	If \mbox{$S \subseteq \mcs$}, then \mbox{$AGRUn(S) \in \mcb$}.
\end{prop}

Not surprisingly, by taking all opinions of all sources seriously, we
may generate many conflicts, manifested as fully connected subsets of
$\mcw$.

\begin{exmp}
	Suppose all three sources in the space robot scenario of
	Example~\ref{ex-robot} are considered equally credible, then
	the aggregate belief state will be the fully connected
	relation indicating that there are conflicts over every
	belief.
\end{exmp}

\subsection{Strictly-ranked sources aggregation}
%	    -----------------------------------

Next, consider the case where the sources are strictly ranked, i.e.,
$\sqsupseteq_S$ is a total order. We define an operator such that
lower-ranked sources refine the belief states of higher ranked
sources. That is, in determining the ordering of a pair of worlds, the
opinions of higher-ranked sources generally override those of
lower-ranked sources, and lower-ranked sources are consulted when
higher-ranked sources are agnostic:
\begin{defn}\label{def-AGRRf}
	If \mbox{$S \subseteq \mcs$}, then \mbox{$AGRRf(S)$} is the
	relation \\
	\centerline{\mbox{$\left\{(x, y) : \exists s \in S.\ x <^s y
		\wedge \left(\forall s' \sqsupset s \in S.\ x
		\sagnost^{s'} y\right)\right\}$}.}
\end{defn}
The definition of the \mbox{$AGRRf$} operator does not rely on
$\sqsupseteq_S$ being a total order, and we will use it in this more
general setting in the following sub-section. However, in the case
that $\sqsupseteq_S$ is a total order, the result of applying
\mbox{$AGRRf$} is guaranteed to be a belief state.
\begin{prop}\label{prop-AGRRf}
	If \mbox{$S \subseteq \mcs$} and $\sqsupseteq_S$ is a total
	order, then \mbox{$AGRRf(S) \in \mcb$}.
\end{prop}

\begin{exmp}\label{ex-robot-AGRRf}
	Suppose, in the space robot scenario of
	Example~\ref{ex-robot}, the technician is considered more
	credible than the manager who, in turn, is considered more
	credible than the programmer. The aggregate belief state,
	shown in Figure~\ref{fig-robot-AGRRf}, informs the robot
	correctly that the feedback system has crashed, but that it
	shouldn't worry about an overload problem and should keep
	sending data.

	\begin{figure}[htb]
	\centerline{\resizebox{!}{2in}{\includegraphics{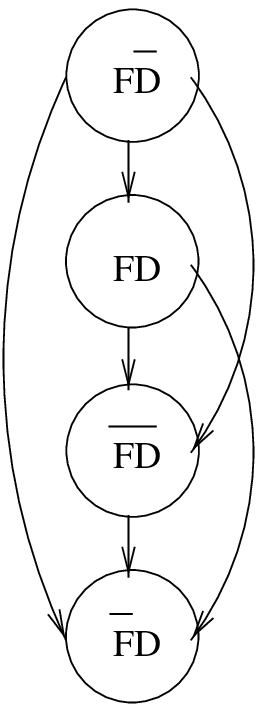}}}
	%\centerline{\epsfig{file=robot-AGRRf.eps,height=2in}}
	\caption{The belief state after aggregation in
	Example~\ref{ex-robot-AGRRf} when \mbox{$s_t \sqsupset s_m
	\sqsupset s_p$}.}\label{fig-robot-AGRRf}
	\end{figure}
\end{exmp}

Note that this case of strictly-ranked sources is almost exactly that
considered in~\cite{MS00}, except that the authors are not able to
allow for conflicts in belief states. A surprising result they show is
that standard AGM belief revision~\cite{AGM85} can be modeled as
the aggregation of two sources, the informant and the informee, where
the informant is considered more credible than the informee.

\subsection{General aggregation}
%	    -------------------

In the general case, we may have several ranks represented and
multiple sources of each rank. It will be instructive to first
consider the following seemingly natural strawman operator, \mbox{$AGR^*$}:
First combine equi-rank sources using \mbox{$AGRUn$}, then aggregate the
strictly-ranked results using what is essentially \mbox{$AGRRf$}:
\begin{defn}
	Let \mbox{$S \subseteq \mcs$}. For any $r \in \mcr$, let
	\mbox{$<_r = AGRUn(\{s \in S: rank(s) = r\})$} and
	$\sagnost_{r'}$, the corresponding agnosticism relation. Also,
	let \mbox{$rank(S) = \{r \in \mcr : \exists s \in S.\ rank(s)
	= r\}$}. \mbox{$AGR^*(S)$} is the relation
	\begin{displaymath}
		\left\{(x, y) : \begin{array}{l}
			\exists r \in \mcr.\ x <_r y \wedge \\
			\left(\forall r' > r \in ranks(S).\ x
				\sagnost_{r'} y\right)
			\end{array}
		\right\}
	\end{displaymath}
\end{defn}
$AGR^*$ indeed defines a legitimate belief state:
\begin{prop}
	If \mbox{$S \subseteq \mcs$}, then \mbox{$AGR^*(S) \in \mcb$}.
\end{prop}

Unfortunately, a problem with this ``divide-and-conquer'' approach is
it assumes the result of aggregation is independent of potential
interactions between the individual sources of different
ranks. Consequently, opinions that will eventually get overridden may
still have an indirect effect on the final aggregation result by
introducing superfluous opinions during the intermediate equi-rank
aggregation step, as the following example shows:
\begin{exmp}\label{ex-AGR*}
	Let \mbox{$\mcw = \{a, b, c\}$}. Suppose \mbox{$S \subseteq
	\mcs$} such that \mbox{$S = \{s_0, s_1, s_2\}$} with belief
	states \mbox{$<^{s_0} = \{(b, a), (b, c)\}$} and
	\mbox{$<^{s_1} = <^{s_2} = \{(a, b), (c, b)\}$}, and where
	\mbox{$s_2 \sqsupset s_1 \sqeq s_0$}. Then \mbox{$AGR^*(S)$}
	is \mbox{$\{(a, b), (c, b), (a, c), (c, a), (a, a), (b, b),
	(c, c)\}$}. All sources are agnostic over $a$ and $c$, yet
	\mbox{$(a, c)$} and \mbox{$(c, a)$} are in the result because
	of the transitive closure in the lower rank involving opinions
	(\mbox{$(b, c)$} and \mbox{$(b, a)$}) which actually get
	overridden in the final result.
\end{exmp}

Because of these undesired effects, we propose another aggregation
operator which circumvents this problem by applying refinement (as
defined in Definition~\ref{def-AGRRf}) to the set of source belief
states before infering new opinions via closure:
\begin{defn}\label{def-AGR}
	The {\em rank-based aggregation} of a set of sources \mbox{$S
	\subseteq \mcs$} is \mbox{$AGR(S) = AGRRf(S)^+$}.
\end{defn}
Encouragingly, \mbox{$AGR$} outputs a valid belief state:
\begin{prop}\label{prop-AGR}
	If \mbox{$S \subseteq \mcs$}, then \mbox{$AGR(S) \in \mcb$}.
\end{prop}

\begin{exmp}\label{ex-robot-AGR}
	Suppose, in the space robot scenario of
	Example~\ref{ex-robot}, the technician is still considered
	more credible than the manager and the programmer, but the
	latter two are considered equally credible. The aggregate
	belief state, shown in Figure~\ref{ex-robot-AGR}, still gives
	the robot the correct information about the state of the
	system. The robot also learns for future reference that there
	is some disagreement over whether or not there would have been
	a data overload if the feedback system were working.

	\begin{figure}[htb]
	\centerline{\resizebox{!}{1.5in}{\includegraphics{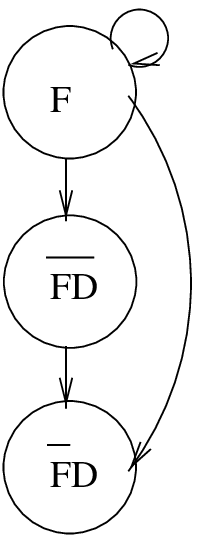}}}
	%\centerline{\epsfig{file=robot-AGR.eps,height=1.5in}}
	\caption{The belief state after aggregation in
	Example~\ref{ex-robot-AGR} when \mbox{$s_t \sqsupset s_m
	\sqeq s_p$}.}\label{fig-robot-AGR}
	\end{figure}
\end{exmp}

We observe that \mbox{$AGR$}, when applied to the set of sources in
Example~\ref{ex-AGR*}, does indeed bypass the problem described above
of extraneous opinion introduction:
\begin{exmp}
	Assume $\mcw$, $S$, and $\sqsupseteq$ are as in
	Example~\ref{ex-AGR*}. \mbox{$AGR(S) = \{(a, b), (c, b)\}$}.
\end{exmp}
We also observe that \mbox{$AGR$} behaves well in the special cases we've
considered, reducing to \mbox{$AGRUn$} when all sources have equal rank, and
to \mbox{$AGRRf$} when the sources are totally ranked:
\begin{prop}\label{prop-AGRSpecial}
	Suppose \mbox{$S \subseteq \mcs$}.
	\begin{enumerate}

	\item If $\sqsupseteq_S$ is fully connected, \mbox{$AGR(S) =
	AGRUn(S)$}.

	\item If \mbox{$\sqsupseteq_S$} is a total order,
	\mbox{$AGR(S) = AGRRf(S)$}.

	\end{enumerate}
\end{prop}

\subsection{Arrow, revisited}
%	    ----------------

Finally, a strong argument in favor of \mbox{$AGR$} is that it
satisfies appropriate modifications of Arrow's conditions. Let $f$ be
an operator which aggregates the belief states $<^{s_1}$, \ldots,
$<^{s_n}$ over $\mcw$ of $n$ sources \mbox{$s_1, \dots, s_n \in
\mcs$}, respectively, and let \mbox{$\prec\ = f(<^{s_1}, \ldots,
<^{s_n})$}. We consider each condition separately.

\paragraph{Restricted range}
The output of the aggregation function will be a modular, transitive
belief state rather than a total pre-order.

\begin{defn}
	{\em (modified) Restricted Range}: The range of $f$ is $\mcb$.
\end{defn}

\paragraph{Unrestricted domain}
Similarly, the input to the aggregation function will be modular,
transitive belief states of sources rather than total pre-orders.

\begin{defn}
	{\em (modified) Unrestricted Domain}: For each $i$, $<^{s_i}$
	can be any member of $\mcb$.
\end{defn}

\paragraph{Pareto principle}
Generalized belief states already represent strict
likelihood. Consequently, we use the actual input and output relations
of the aggregation function in place of their strict versions to
define the Pareto principle. Obviously, because we allow for the
introduction of conflicts, \mbox{$AGR$} will not satisfy the original
formal Pareto principle which essentially states that if all sources
have an unconflicted belief that one world is strictly more likely
than another, this must also be true of the aggregated belief
state. Neither condition is necessarily stronger than the other.

\begin{defn}
	{\em (modified) Pareto Principle}: If \mbox{$x <^{s_i} y$} for
	all $i$, then \mbox{$x \prec y$}.
\end{defn}

\paragraph{Independence of irrelevant alternatives}
Conflicts are defined in terms of cycles, not necessarily binary. By
allowing the existence of conflicts, we effectively have made it
possible for outside worlds to affect the relation between a pair of
worlds, viz., by involving them in a cycle. As a result, we need to
weaken IIA to say that the relation between worlds should be
independent of other worlds {\bf unless} these other worlds put them
in conflict.

\begin{defn}
	{\em (modified) Independence of Irrelevant Alternatives
	(IIA)}: Suppose \mbox{$s'_1, \ldots, s'_n \in \mcs$} such that
	\mbox{$s_i \sqeq s'_i$} for all $i$, and \mbox{$\prec' =
	f(<^{s'_1}, \ldots, <^{s'_n})$}. If, for \mbox{$x, y \in
	\mcw$}, \mbox{$x <^{s_i} y$} iff \mbox{$x <^{s'_i} y$} for all
	$i$, \mbox{$x \notconflict y$}, and \mbox{$x \notconflict'
	y$}, then \mbox{$x \prec y$} iff \mbox{$x \prec' y$}.
\end{defn}

\paragraph{Non-dictatorship}
As with the Pareto principle definition, we use the actual input and
output relations to define non-dictatorship since belief states
represent strict likelihood. From this perspective, our setting
requires that informant sources of the highest rank be ``dictators''
in the sense considered by Arrow. However, the setting originally
considered by Arrow was one where all individuals are ranked
equally. Thus, we make this explicit in our new definition of
non-dictatorship by adding the pre-condition that all sources be of
equal rank. Now, \mbox{$AGR$} treats a set of equi-rank sources
equally by taking all their opinions seriously, at the price of
introducing conflicts. So, intuitively, there are no
dictators. However, because Arrow did not account for conflicts in his
formulation, all the sources will be ``dictators'' by his
definition. We need to modify the definition of non-dictatorship to
say that no source can always push opinions through without them ever
being contested.

\begin{defn}
	{\em (modified) Non-Dictatorship}: If \mbox{$s_i \sqeq s_j$}
	for all \mbox{$i, j$}, then there is no $i$ such that, for
	every combination of source belief states and every \mbox{$x,
	y \in \mcw$}, \mbox{$x <^{s_i} y$} and \mbox{$y \not<^{s_i}
	x$} implies \mbox{$x \prec y$} and \mbox{$y \not\prec x$}.
\end{defn}

We now show that \mbox{$AGR$} indeed satisfies these conditions:
\begin{prop}\label{prop-Arrow2}
	Let \mbox{$S = \{s_1, \ldots, s_n\} \subseteq \mcs$} and
	\mbox{$AGR_f(<^{s_1}, \ldots, <^{s_n}) =
	AGR(S)$}. \mbox{$AGR_f$} satisfies (the modified versions of)
	restricted range, unrestricted domain, Pareto principle, IIA,
	and non-dictatorship.
\end{prop}

\section{Multi-agent fusion}
%	 __________________

So far, we have only considered the case where a single agent must
construct or update her belief state once informed by a set of
sources. Multi-agent fusion is the process of aggregating the belief
states of a set of agents, each with its respective set of informant
sources. We proceed to formalize this setting.

An agent $A$ is {\em informed by} a set of sources \mbox{$S \subseteq
\mcs$}. Agent $A$'s {\em induced belief state} is the belief state
formed by aggregating the belief states of its informant sources,
i.e., \mbox{$AGR(S)$}. Assume the set of agents to fuse agree upon
\mbox{$rank$} (and, consequently, $\sqsupseteq$).\footnote{We could
easily extend the framework to allow for individual rankings, but we
felt that the small gain in generality would not justify the
additional complexity and loss of perspicuity. Similarly, we could
consider each agent as having a credibility ordering only over its
informant sources. However, it is unclear how, for example, crediblity
orderings over disjoint sets of sources should be combined into a new
credibility ordering since their union will not be total.}  We define
the fusion of this set to be an agent informed by the combination of
informant sources:
\begin{defn}\label{def-fusion}
	Let \mbox{$\mca = \{A_1, \ldots, A_n\}$} be a set of agents
	such that each agent $A_i$ is informed by \mbox{$S_i \subseteq
	\mcs$}. The {\em fusion} of $\mca$, written
	\mbox{$\fuse(\mca)$}, is an agent informed by \mbox{$S =
	\bigcup^n_{i=1} S_i$}.
\end{defn}
Not surprisingly given its set-theoretic definition, fusion is
idempotent, commutative, and associative. These properties guarantee
the invariance required in multi-agent belief aggregation applications
such as our space robot domain.

In the multi-agent space robot scenario described in
Section~\ref{sec-intro}, we only have a direct need for the belief
states that result from fusion. We are only interested in the belief
states of the original sources in as far as we want the fused belief
state to reflect its informant history. An obvious question is whether
it is possible to compute the belief state induced by the agents'
fusion solely from their initial belief states, that is, without
having to reference the belief states of their informant sources. This
is highly desirable because of the expense of storing---or, as in the
case of our space robot example, transmitting---all source belief
states; we would like to represent each agent's knowledge as compactly
as possible.

In fact, we can do this if all sources have equal rank. We simply take
the transitive closure of the union of the agents' belief states:
\begin{prop}\label{prop-FuseEq}
	Let $\mca$ and $S$ be as in Definition~\ref{def-fusion},
	$\prec^{A_i}$, agent $A_i$'s induced belief state, and
	$\sqsupseteq_S$, fully connected. If \mbox{$A = \fuse(\mca)$},
	then \mbox{$\left(\bigcup_{A_i \in \mca}
	\prec^{A_i}\right)^+$} is $A$'s induced belief state.
\end{prop}

Unfortunately, the equal rank case is special. If we have sources of
different ranks, we generally cannot compute the induced belief state
after fusion using only the agent belief states before fusion, as the
following simple example demonstrates:
\begin{exmp}
	Let \mbox{$\mcw = \{a, b\}$}. Suppose two agents $A_1$ and
	$A_2$ are informed by sources $s_1$ with belief state
	\mbox{$<^{s_1} = \{(a, b)\}$} and $s_2$ with belief state
	\mbox{$<^{s_2} = \{(b, a)\}$}, respectively. $A_1$'s belief
	state is the same as $s_1$'s and $A_2$'s is the same as
	$s_2$'s. If \mbox{$s_1 \sqsupset s_2$}, then the belief state
	induced by \mbox{$\fuse(A_1, A_2)$} is $<^{s_1}$, whereas if
	\mbox{$s_2 \sqsupset s_1$}, then it is $<^{s_2}$. Thus, just
	knowing the belief states of the fused agents is not
	sufficient for computing the induced belief state. We need
	more information about the original sources.
\end{exmp}

However, if sources are totally pre-ordered by credibility, we can
still do much better than storing all the original sources.  It is
enough to store for each opinion of \mbox{$AGRRf(S)$} the rank of the
highest-ranked source supporting it. We define {\em pedigreed belief
states} which enrich belief states with this additional information:
\begin{defn}\label{def-PBS}
	Let $A$ be an agent informed by a set of sources \mbox{$S
	\subseteq \mcs$}. $A$'s {\em pedigreed belief state} is a pair
	$(\prec, l)$ where $\prec = AGRRf(S)$ and $l : \prec \to \mcr$
	such that $l((x,y)) = \max\{rank(s) : x <^s y, s \in S\}$. We
	use $\prec^A_r$ to denote the {\em restriction of $A$'s
	pedigreed belief state to $r$}, that is, \mbox{$\prec^A_r =
	\{(x,y) \in \prec : l((x,y)) = r\}$}.
\end{defn}
We verify that a pair's label is, in fact, the rank of the source used
to determine the pair's membership in \mbox{$AGRRf(S)$}, not that of
some higher-ranked source:
\begin{prop}\label{prop-label}
	Let $A$ be an agent informed by a set of sources \mbox{$S
	\subseteq \mcs$} and with pedigreed belief state
	\mbox{$(\prec, l)$}. Then
	\begin{center}
	\begin{displaymath}
		x \prec^A_r y
	\end{displaymath}
	iff
	\begin{displaymath}
		\begin{array}{l}
		\exists s \in S.\ x <^s y \wedge r = rank(s) \wedge \\
		\left(\forall s' \sqsupset s \in S.\ x \sagnost^{s'}
			y\right).
		\end{array}
	\end{displaymath}
	\end{center}
\end{prop}
The belief state induced by a pedigreed belief state \mbox{$(\prec,
l)$} is, obviously, the transitive closure of $\prec$.

Now, given only the pedigreed belief states of a set of agents, we can
compute the new pedigreed belief state after fusion. We simply combine
the labeled opinions using our refinement techniques.
\begin{prop}
	Let $\mca$ and $S$ be as in Definition~\ref{def-fusion},
	$\sqsupseteq_S$, a total pre-order, and \mbox{$A =
	\fuse(\mca)$}. If
	\begin{enumerate}

	\item $\prec$ is the relation
	\begin{displaymath}
		\left\{(x,y) :
		\begin{array}{l}
			\exists A_i \in \mca, r \in \mcr.\ x
				\prec^{A_i}_r y \wedge \\
			\left(\forall A_j \in \mca, r' > r \in
				\mcr.\ x \agnost^{A_j}_{r'} y\right)
		\end{array}
		\right\} 
	\end{displaymath}
	over $\mcw$,

	\item \mbox{$l : \prec \to \mcr$} such that \mbox{$l((x,y)) =
	\max\{r : x \prec^{A_i}_r y, A_i \in \mca\}$}, and

	\end{enumerate}
	then \mbox{$(\prec, l)$} is $A$'s pedigreed belief state.
\end{prop}
From the perspective of the induced belief states, we are essentially
discarding unlabeled opinions (i.e., those derived by the closure
operation) before fusion. Intuitively, we are learning new information
so we may need to retract some of our inferred opinions. After fusion,
we re-apply closure to complete the new belief state. Interestingly,
in the special case where the sources are strictly-ranked, the closure
is unnecessary:
\begin{prop}
	If $\mca$ and $S$ are as in Definition~\ref{def-fusion},
	$\sqsupseteq_S$ is a total order, and \mbox{$(\prec, l)$} is
	the pedigreed belief state of \mbox{$\fuse(\mca)$}, then
	\mbox{$\prec^+ = \prec$}.
\end{prop}

\begin{exmp}\label{ex-robot-fusion}
	Let's look once more at the space robot scenario considered in
	Example~\ref{ex-robot}. Suppose the arrogant programmer is not
	part of the telemetry team, but instead works for a company on
	the other side of the country.  Then the robot has to request
	information from two separate agents, one to query the manager
	and technician and one to query the programmer. Assume that
	the agents and the robot all rank the sources the same,
	assigning the technician rank 2 and the other two agents rank
	1, which induces the same credibility ordering used in
	Example~\ref{ex-robot-AGR}. The agents' pedigreed belief
	states and the result of their fusion are shown in
	Figure~\ref{fig-robot-fusion}.

	\begin{figure}[htb]
	\centerline{\resizebox{!}{2in}{\includegraphics{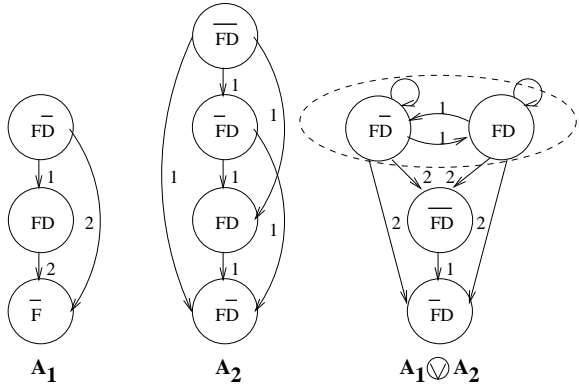}}}
	%\centerline{\epsfig{file=robot-fusion.eps,height=2in}}
	\caption{The pedigreed belief states of agent $A_1$
	informed by $s_m$ and $s_t$ and of agent $A_2$ informed by
	$s_p$, and the result of their fusion in
	Example~\ref{ex-robot-fusion}.}\label{fig-robot-fusion}
	\end{figure}

	The first agent does not provide any information about
	overloading and the second agent provides incorrect
	information. However, we see that after fusing the two, the
	robot has a belief state that is identical to what it computed
	in Example~\ref{ex-robot-AGR} when there was only one agent
	informed by all three sources (we've only separated the top
	set of worlds so as to show the labeling). Consequently, it
	now knows the correct state of the system. And, satisfyingly,
	the final result does not depend on the order in which the
	robot receives the agents' reports.
\end{exmp}

The savings obtained in required storage space by this scheme can be
substantial. Whereas explicitly storing all of an agent's informant
sources $S$ requires \mbox{$\mathcal{O}(\|S\|2^\mcw)$} amount of space
in the worst case (when all the sources' belief states are fully
connected relations), storing a pedigreed belief state only requires
\mbox{$\mathcal{O}(2^\mcw)$} space in the worst case. Moreover, not
only does the enriched representation allow us to conserve space, but
it also provides for potential savings in the efficiency of computing
fusion since, for each pair of worlds, we only need to consider the
opinions of the {\em agents} rather than those of all the sources in
the combined set of informants.

Incidentally, if we had used \mbox{$AGR^*$} as the basis for our general
aggregation, simply storing the rank of the maximum supporting sources
would not give us sufficient information to compute the induced belief
state after fusion. To demonstrate this, we give an example where two
pairs of sources induce the same annotated agent belief states, yet
yield different belief states after fusion:
\begin{exmp}
	Let $\mcw$, $\mcs$, and $\sqsupseteq$ be as in
	Example~\ref{ex-AGR*}. Suppose agents $A_1$, $A_2$, $A_1'$,
	and $A_2'$ are informed by sets of sources $S_1$, $S_2$,
	$S_1'$, and $S_2'$, respectively, where \mbox{$S_1 = S_2 =
	\{s_2\}$}, \mbox{$S_1' = \{s_0, s_2\}$}, and \mbox{$S_2' =
	\{s_1, s_2\}$}. \mbox{$AGR^*$} dictates that the pedigreed
	belief states of all four agents equal $<^{s_2}$ with all
	opinions annotated with \mbox{$rank(s_2)$}. In spite of this
	indistinguishability, if \mbox{$A = \fuse(\{A_1, A_2\})$} and
	\mbox{$A' = \fuse(\{A_1', A_2'\})$}, then $A$'s induced belief
	state equals $<^{s_2}$, i.e., \mbox{$\{(a, b), (c, b)\}$},
	whereas $A'$'s is \mbox{$\{(a, b), (c, b), (a, c), (c, a), (a,
	a), (b, b), (c, c)\}$}.
\end{exmp}

\section{Conclusion}
%	 __________

We have described a semantically clean representation for aggregate
beliefs which allows us to represent conflicting opinions without
sacrificing the ability to make decisions. We have proposed an
intuitive operator which takes advantage of this representation so
that an agent can combine the belief states of a set of informant
sources totally pre-ordered by credibility. Finally, we have described
a mechanism for fusing the belief states of different agents which
iterates well.

The aggregation methods we have discussed here are just special cases
of a more general framework based on voting. That is, we account not
only for the ranking of the sources supporting or disagreeing with an
opinion (i.e., the {\em quality} of support), but also the percentage
of sources in each camp (the {\em quantity} of support). Such an
extension allows for a much more refined approach to aggregation, one
much closer to what humans often use in practice. Exploring this
richer space is the subject of further research.

Another problem which deserves further study is developing a fuller
understanding of the properties of the \mbox{$Bel$}, \mbox{$Agn$}, and
\mbox{$Con$} operators and how they interrelate.

\subsubsection*{Acknowledgements}
%		________________

Pedrito Maynard-Reid II was partly supported by a National Physical
Science Consortium Fellowship.  The final version of this paper was
written with the financial support of the Jean et H\'{e}l\`{e}ne
Alfassa Fund for Research in Artificial Intelligence.

\bibliography{mtfuse}

\begin{thebibliography}{}

\bibitem[\protect\citeauthoryear{Alchourr\'{o}n \bgroup \em et al.\egroup
  }{1985}]{AGM85}
Carlos~E. Alchourr\'{o}n, Peter G\"{a}rdenfors, and David Makinson.
\newblock On the logic of theory change: Partial meet contraction and revision
  functions.
\newblock {\em Journal of Symbolic Logic}, 50:510--530, 1985.

\bibitem[\protect\citeauthoryear{Arrow}{1963}]{Arrow63}
Kenneth~J. Arrow.
\newblock {\em Social Choice and Individual Values}.
\newblock Wiley, New York, 2nd edition, 1963.

\bibitem[\protect\citeauthoryear{G\"{a}rdenfors and Makinson}{1994}]{GM94}
Peter G\"{a}rdenfors and David Makinson.
\newblock Nonmonotonic inference based on expectations.
\newblock {\em Artificial Intelligence}, 65(1):197--245, January 1994.

\bibitem[\protect\citeauthoryear{Grove}{1988}]{Grove88}
Adam Grove.
\newblock Two modellings for theory change.
\newblock {\em Journal of Philosophical Logic}, 17:157--170, 1988.

\bibitem[\protect\citeauthoryear{Kahneman and Tversky}{1979}]{KT79}
D.~Kahneman and A.~Tversky.
\newblock Prospect theory: An analysis of decision under risk.
\newblock {\em Econometrica}, 47(2):263--291, March 1979.

\bibitem[\protect\citeauthoryear{Katsuno and Mendelzon}{1991}]{KM91}
Hirofumi Katsuno and Alberto~O. Mendelzon.
\newblock Propositional knowledge base revision and minimal change.
\newblock {\em Artificial Intelligence}, 52(3):263--294, 1991.

\bibitem[\protect\citeauthoryear{Kreps}{1990}]{Kreps90}
David~M. Kreps.
\newblock {\em A Course in Microeconomic Theory}.
\newblock Princeton University Press, 1990.

\bibitem[\protect\citeauthoryear{Lehmann and Magidor}{1992}]{LM92}
Daniel Lehmann and Menachem Magidor.
\newblock What does a conditional knowledge base entail?
\newblock {\em Artificial Intelligence}, 55(1):1--60, May 1992.

\bibitem[\protect\citeauthoryear{{Maynard-Reid II} and Shoham}{2000}]{MS00}
Pedrito {Maynard-Reid II} and Yoav Shoham.
\newblock Belief fusion: Aggregating pedigreed belief states.
\newblock {\em Journal of Logic, Language, and Information}, 2000.
\newblock To appear.

\bibitem[\protect\citeauthoryear{Sen}{1986}]{Sen86}
Amartya Sen.
\newblock Social choice theory.
\newblock In K.~J. Arrow and M.~D. Intriligator, editors, {\em Handbook of
  Mathematical Economics}, volume III, chapter~22, pages 1073--1181. Elevier
  Science Publishers, 1986.

\end{thebibliography}

\newpage

\begin{appendix}

\section{Proofs}
%	 ______

\setcounter{prop}{0}

\begin{prop}\hspace{0ex}\\[-0.5cm]
	\begin{enumerate}

	\item The transitive closure of a modular relation is
	modular.

	\item Every transitive relation is quasi-transitive.

	\item \cite{Sen86} Every quasi-transitive relation is
	acyclic.

	\end{enumerate}
\end{prop}
\begin{pf}\hspace{0ex}\\[-0.5cm]
\begin{enumerate}

\item Suppose a relation $\leq$ over finite set $\Omega$ is modular,
and $\leq^+$ is the transitive closure of $\leq$. Suppose \mbox{$x, y,
z \in \Omega$} and \mbox{$x \leq^+ y$}. Then there exist \mbox{$w_0,
\ldots, w_n$} such that \mbox{$x = w_0 \leq \cdots \leq w_n =
y$}. Since $\leq$ is modular and \mbox{$w_0 \leq w_1$}, either
\mbox{$w_0 \leq z$} or \mbox{$z \leq w_1$}. In the former case,
\mbox{$x = w_0 \leq z$}, so \mbox{$x \leq^+ z$}. In the latter case,
\mbox{$z \leq w_1 \leq \cdots w_n = y$}, so \mbox{$z \leq^+ y$}.

\item Suppose $\Omega$ is a finite set, \mbox{$x, y, z \in \Omega$},
$\leq$ is a transitive relation over $\Omega$, and $<$ is its strict
version. Suppose \mbox{$x < y$} and \mbox{$y < z$}. Then \mbox{$x \leq
y$}, \mbox{$y \not\leq x$}, \mbox{$y \leq z$}, and \mbox{$z \not\leq
y$}. \mbox{$x \leq y$} and \mbox{$y \leq z$} imply \mbox{$x \leq z$},
and \mbox{$y \leq z$} and \mbox{$y \not\leq x$} imply \mbox{$z
\not\leq x$}, both by transitivity. So \mbox{$x < z$}.

\end{enumerate}
\qed
\end{pf}

\begin{prop}
	\cite{Sen86} Given a relation $\leq$ over a finite set
	$\Omega$, the choice set operation $\C$ defines a choice
	function iff $\leq$ is acyclic.
\end{prop}
\begin{pf}
See~\cite{Sen86}.
\qed
\end{pf}

\begin{prop}
	\cite{Arrow63} There is no aggregation operator that satisfies
	restricted range, unrestricted domain, (weak) Pareto
	principle, independendence of irrelevant alternatives, and
	nondictatorship.
\end{prop}
\begin{pf}
See~\cite{Arrow63}.
\qed
\end{pf}

\begin{prop}
	Let $\preceq$ be a relation over a finite set $\Omega$ and let
	$\agnost$ be its symmetric restriction (i.e., \mbox{$x \agnost
	y$} iff \mbox{$x \preceq y$} and \mbox{$y \preceq x$}). If
	$\preceq$ is total and quasi-transitive but not transitive,
	then $\agnost$ is not transitive.
\end{prop}
\begin{pf}
Let $\preceq$ be a total, quasi-transitive, non-transitive
relation. First, such a relation exits: if \mbox{$\Omega = \{a, b,
c\}$}, it is easily verified that the relation \mbox{$\Omega \times
\Omega \setminus \{(b,a)\}$} is total, quasi-transitive, but not
transitive.

Suppose \mbox{$x \preceq y$} and \mbox{$y \preceq z$} but \mbox{$x
\not\preceq z$}. By totality, \mbox{$z \preceq x$}, so \mbox{$z \prec
x$}. If \mbox{$x \prec y$}, then \mbox{$z \prec y$} by
quasi-transitivity, a contradiction. Thus, \mbox{$x \agnost
y$}. Similarly, if \mbox{$y \prec z$}, then \mbox{$y \prec x$}, a
contradiction, so \mbox{$y \agnost z$}. But \mbox{$z \prec x$}, so
\mbox{$x \not\agnost z$}. Therefore, $\agnost$ is not transitive.
\qed
\end{pf}

\begin{prop}
	Suppose a relation $\prec$ is transitive and $\agnost$ is the
	corresponding agnosticism relation. Then $\agnost$ is
	transitive iff $\prec$ is modular.
\end{prop}
\begin{pf}
Suppose $\agnost$ is transitive and suppose \mbox{$x \prec z$},
\mbox{$x, y, z \in \mcw$}. We prove by contradiction: Suppose \mbox{$x
\not\prec y$} and \mbox{$y \not\prec z$}. By transitivity, \mbox{$z
\not\prec y$} and \mbox{$y \not\prec x$}, so \mbox{$x \agnost y$} and
\mbox{$y \agnost z$}. By assumption, \mbox{$x \agnost z$}, so \mbox{$x
\not\prec z$}, a contradiction.

Suppose, instead, $\prec$ is modular and suppose \mbox{$x \agnost y$}
and \mbox{$y \agnost z$}, \mbox{$x, y, z \in \mcw$}. Then \mbox{$x
\not\prec y$}, \mbox{$y \not\prec x$}, \mbox{$y \not\prec z$}, and
\mbox{$z \not\prec y$}. By modularity, \mbox{$x \not\prec z$} and
\mbox{$z \not\prec x$}, so \mbox{$x \agnost z$}.
\qed
\end{pf}

\begin{prop}
	The set of irreflexive relations in $\mcb$ is isomorphic to
	$\mct$ and, in fact, equals $\mcst$.
\end{prop}
\begin{pf}
Let \mbox{$x, y, z \in \mcw$}.  Suppose \mbox{$\prec \in \mcb$} is
irreflexive. Let $\preceq$ be defined as \mbox{$x \preceq y$} iff
\mbox{$y \not\prec x$}.  We first show that $\prec$ is the strict
version of $\preceq$. Suppose $\prec'$ is the strict version of
$\preceq$. If \mbox{$x \prec' y$}, then \mbox{$x \preceq y$} and
\mbox{$y \not\preceq x$}, so \mbox{$x \prec y$}. If, instead, \mbox{$x
\prec y$}, then \mbox{$y \not\preceq x$}. By totality, \mbox{$x
\preceq y$}, so \mbox{$x \prec' y$}.

We show that \mbox{$\preceq \in \mct$}. If \mbox{$x \not\prec y$} then
\mbox{$y \preceq x$}. Otherwise, \mbox{$x \prec y$}. But since $\prec$
is irreflexive, \mbox{$y \not\prec x$} (otherwise \mbox{$x \prec x$}
by transitivity), so \mbox{$x \preceq y$} and $\preceq$ is total.
Next, suppose \mbox{$x \preceq y$} and \mbox{$y \preceq z$}. Then
\mbox{$y \not\prec x$} and \mbox{$z \not\prec y$}. By modularity,
\mbox{$z \not\prec x$}, so \mbox{$x \preceq z$}, so $\preceq$ is
transitive.

Now suppose \mbox{$\preceq \in \mct$} and $\prec$ is its strict
version. First we show that $\prec$ is modular. Suppose \mbox{$x \prec
y$}. Then \mbox{$x \preceq y$} and \mbox{$y \not\preceq x$}. Since
$\preceq$ is total, \mbox{$x \preceq z$} or \mbox{$z \preceq
x$}. Suppose \mbox{$x \preceq z$}. Whether \mbox{$y \preceq z$} or
\mbox{$y \not\preceq z$}, \mbox{$z \not\preceq x$} by
transitivity. Suppose, instead, \mbox{$z \preceq x$}. Then \mbox{$z
\preceq y$} and \mbox{$y \not\preceq z$}, both by transitivity. We
conclude that \mbox{$x \preceq z$} and \mbox{$z \not\preceq x$}, or
\mbox{$z \preceq y$} and \mbox{$y \not\preceq z$}, so \mbox{$x \prec
z$} or \mbox{$z \prec y$}.  Second, transitivity of $\prec$ follows
immediately from Proposition~\ref{prop-Props} and the transitivity of
$\preceq$. Finally, $\prec$ is irreflexive since it is asymmetric.
\qed
\end{pf}

\begin{prop}
	\mbox{$\prec \in \mcb$} iff there is a partition
	\mbox{$\mathbf{W} = \langle W_0, \ldots, W_n \rangle$} of
	$\mcw$ such that:
	\begin{enumerate}

	\item For every \mbox{$x \in W_i$} and \mbox{$y \in W_j$},
	\mbox{$i \neq j$} implies \mbox{$i < j$} iff \mbox{$x \prec
	y$}.

	\item Every $W_i$ is either fully connected (\mbox{$w \prec
	w'$} for all \mbox{$w, w' \in W_i$}) or fully disconnected
	(\mbox{$w \not\prec w'$} for all \mbox{$w, w' \in W_i$}).

	\end{enumerate}
\end{prop}
\begin{pf}
We refer to the conditions in the proposition as conditions~1 and 2,
respectively. We prove each direction of the proposition separately.

$(\Longrightarrow)$ Suppose \mbox{$\prec \in \mcb$}, that is, $\prec$
is a modular and transitive relation over $\mcw$. We use a series of
definitions and lemmas to show that a partition of $\mcw$ exists
satisfying conditions~1 and 2. We first define an equivalence relation
by which we will partition $\mcw$. Two elements will be equivalent if
they ``look the same'' from the perspective of every element of
$\mcw$:
\begin{defn}
	\mbox{$x \equiv y$} iff for every \mbox{$z \in \mcw$},
	\mbox{$x \prec z$} iff \mbox{$y \prec z$} and \mbox{$z \prec
	x$} iff \mbox{$z \prec y$}.
\end{defn}

\begin{lem}
	$\equiv$ is an equivalence relation over $\mcw$.
\end{lem}
\begin{pf}
Suppose \mbox{$x \in \mcw$}. For every \mbox{$z \in \mcw$}, \mbox{$x
\prec z$} iff \mbox{$x \prec z$} and \mbox{$z \prec x$} iff \mbox{$z
\prec x$}, so \mbox{$x \equiv x$}. Therefore, $\equiv$ is reflexive.

Suppose \mbox{$x, y \in \mcw$} and \mbox{$x \equiv y$}. Then for every
\mbox{$z \in \mcw$}, \mbox{$x \prec z$} iff \mbox{$y \prec z$} and
\mbox{$z \prec x$} iff \mbox{$z \prec y$}. But then for every \mbox{$z
\in \mcw$}, \mbox{$y \prec z$} iff \mbox{$x \prec z$} and \mbox{$z
\prec y$} iff \mbox{$z \prec x$}. Therefore, \mbox{$y \equiv x$}, so
$\equiv$ is symmetric.

Suppose \mbox{$x, y, z \in \mcw$}, \mbox{$x \equiv y$}, and \mbox{$y
\equiv z$}. Suppose further that \mbox{$w \in \mcw$}. By definition of
$\equiv$, \mbox{$x \prec w$} iff \mbox{$y \prec w$} and \mbox{$w \prec
x$} iff \mbox{$w \prec y$}, and \mbox{$y \prec w$} iff \mbox{$z \prec
w$} and \mbox{$w \prec y$} iff \mbox{$w \prec z$}. Therefore, \mbox{$x
\prec w$} iff \mbox{$z \prec w$} and \mbox{$w \prec x$} iff \mbox{$w
\prec z$}. Since $w$ is arbitrary, \mbox{$x \equiv z$}, so $\equiv$ is
transitive.
\qed
\end{pf}
$\equiv$ partitions $\mcw$ into its equivalence classes. We use
\mbox{$[w]$} to denote the equivalence class containing $w$, that is,
the set \mbox{$\{w' \in \mcw : w \equiv w'\}$}. Observe that two
worlds in conflict always appear in the same equivalence class:
\begin{lem}\label{lem-conflict-is-equiv}
	If \mbox{$x, y \in \mcw$} and \mbox{$x \conflict y$}, then
	\mbox{$[x] = [y]$}.
\end{lem}
\begin{pf}
Suppose \mbox{$x, y \in \mcw$} and \mbox{$x \conflict y$}. Since
\mbox{$[x]$} is an equivalence class, it suffices to show that
\mbox{$y \in [x]$}, that is, \mbox{$x \equiv y$}. Suppose \mbox{$z \in
\mcw$}. By transitivity, if \mbox{$x \prec z$}, then \mbox{$y \prec
z$}; if \mbox{$y \prec z$}, then \mbox{$x \prec z$}; if \mbox{$z \prec
x$}, then \mbox{$z \prec y$}; and, if \mbox{$z \prec y$} then \mbox{$z
\prec x$}. Thus, \mbox{$x \prec z$} iff \mbox{$y \prec z$} and
\mbox{$z \prec x$} iff \mbox{$z \prec y$}, and since $z$ is arbitrary,
\mbox{$x \equiv y$}.
\qed
\end{pf}

We now define a total order over these equivalence classes:
\begin{defn}
	For all \mbox{$x, y \in \mcw$}, \mbox{$[x] \leq [y]$} iff
	\mbox{$[x] = [y]$} or \mbox{$x \prec y$}.
\end{defn}

\begin{lem}
	$\leq$ is well-defined, that is, if \mbox{$x \equiv x'$} and
	\mbox{$y \equiv y'$}, then \mbox{$x \prec y$} iff \mbox{$x'
	\prec y'$}, for all \mbox{$x, x', y, y' \in \mcw$}.
\end{lem}
\begin{pf}
Suppose \mbox{$x \equiv x'$} and \mbox{$y \equiv y'$}, \mbox{$x, x',
y, y' \in \mcw$}. By the definition of $\equiv$, for every \mbox{$z
\in \mcw$}, \mbox{$x \prec z$} iff \mbox{$x' \prec z$}. In particular,
\mbox{$x \prec y$} iff \mbox{$x' \prec y$}. Also by the definition of
$\equiv$, for every \mbox{$z' \in \mcw$}, \mbox{$z' \prec y$} iff
\mbox{$z' \prec y'$}. In particular, \mbox{$x' \prec y$} iff \mbox{$x'
\prec y'$}. Therefore, \mbox{$x \prec y$} iff \mbox{$x' \prec y'$}.
\qed
\end{pf}

\begin{lem}
	$\leq$ is a total order over the equivalence classes of $\mcw$
	defined by $\equiv$.
\end{lem}
\begin{pf}
Suppose \mbox{$x, y, z \in \mcw$}. We first show that $\leq$ is
total. By definition of $\leq$, if \mbox{$x \prec y$} or \mbox{$y
\prec x$}, then \mbox{$[x] \leq [y]$} or \mbox{$[y] \leq [x]$},
respectively. Suppose \mbox{$x \not\prec y$} and \mbox{$y \not\prec
x$}, and suppose \mbox{$z \in \mcw$}. By modularity of $\prec$,
\mbox{$x \prec z$} implies \mbox{$y \prec z$}, \mbox{$y \prec z$}
implies \mbox{$x \prec z$}, \mbox{$z \prec x$} implies \mbox{$z \prec
y$}, and \mbox{$z \prec y$} implies \mbox{$z \prec x$}, so \mbox{$x
\equiv y$}. Therefore, \mbox{$[x] = [y]$}, so \mbox{$[x] \leq [y]$} by
the definition of $\leq$.

Next, we show that $\leq$ is anti-symmetric. Suppose \mbox{$[x] \leq
[y]$} and \mbox{$[y] \leq [x]$}. Then \mbox{$[x] = [y]$} or \mbox{$x
\prec y$} and \mbox{$y \prec x$}. In the former case we are done, in
the latter, the result follows from Lemma~\ref{lem-conflict-is-equiv}.

Finally, we show that $\leq$ is transitive. Suppose \mbox{$[x] \leq
[y]$} and \mbox{$[y] \leq [z]$}. Obviously, if \mbox{$[x] = [y]$} or
\mbox{$[y] = [x]$}, then \mbox{$[x] \leq [z]$}. Suppose not. Then
\mbox{$x \prec y$} and \mbox{$y \prec z$}, so \mbox{$x \prec z$} by
the transitivity of $\prec$. Therefore, \mbox{$[x] \leq [y]$} by the
definition of $\leq$.
\qed
\end{pf}

We name the members of the partition \mbox{$W_0, \ldots, W_n$} such
that \mbox{$W_i \leq W_j$} iff \mbox{$i \leq j$}, where $n$ is an
integer.  Such a naming exists since every finite, totally ordered set
is isomorphic to some finite prefix of the integers.

We now check that this partition satisfies the two conditions. For the
first condition, suppose \mbox{$x \in W_i$}, \mbox{$y \in W_j$}, and
\mbox{$i \neq j$}. We want to show that \mbox{$i < j$} iff \mbox{$x
\prec y$}. Since \mbox{$i \neq j$}, \mbox{$[x] \neq [y]$}. Suppose
\mbox{$i < j$}. Then \mbox{$i \leq j$}, so \mbox{$[x] \leq
[y]$}. Since \mbox{$[x] \neq [y]$}, \mbox{$x \prec y$} by the
definition of $\leq$. Now suppose, instead, that \mbox{$x \prec
y$}. Then \mbox{$[x] \leq [y]$} by the definition of $\leq$, so
\mbox{$i \leq j$}. Since \mbox{$[x] \neq [y]$}, \mbox{$y \not\prec x$}
by Lemma~\ref{lem-conflict-is-equiv}. Since \mbox{$[x] \neq [y]$} and
\mbox{$y \not\equiv x$}, \mbox{$[x] \not\leq [y]$} by the definition
of $\leq$, so \mbox{$j \not\leq i$}. Thus, \mbox{$i < j$}.

Finally, we show that each $W_i$ is either fully connected or fully
disconnected. Suppose \mbox{$x, y, z \in W_i$} so that \mbox{$x \equiv
y \equiv z$}. It suffices to show that \mbox{$x \prec x$} iff \mbox{$y
\prec z$}. By the definition of $\equiv$, \mbox{$x \prec x$} iff
\mbox{$y \prec x$}, and \mbox{$x \prec x$} iff \mbox{$x \prec
z$}. Suppose \mbox{$x \prec x$}. Then, \mbox{$y \prec x$} and \mbox{$x
\prec z$}, so \mbox{$y \prec z$} by transitivity of $\prec$.  Suppose
now, \mbox{$x \not\prec x$}.  Then, \mbox{$y \not\prec x$} and
\mbox{$x \not\prec z$}, so \mbox{$y \not\prec z$} by modularity of
$\prec$.

$(\Longleftarrow)$ Suppose \mbox{$\mathbf{W} = \langle W_0, \ldots,
W_n \rangle$} is a partition of $\mcw$ and $\prec$ is a relation over
$\mcw$ satisfying the given conditions. We want to show that $\prec$
is modular and transitive. We first give the following lemma:
\begin{lem}\label{lem-prec-imp-leq}
	Suppose $\mathbf{W}$ is a partition of $\mcw$ and $\prec$ is a
	relation over $\mcw$ satisfying condition~1. If \mbox{$W_i,
	W_j \in \mathbf{W}$}, \mbox{$x \in W_i$}, \mbox{$y \in W_j$},
	and \mbox{$x \prec y$}, then \mbox{$i \leq j$}.
\end{lem}
\begin{pf}
If \mbox{$i = j$}, we're done. Suppose \mbox{$i \neq j$}. Then, since
\mbox{$x \prec y$}, \mbox{$i < j$} by condition~1.
\qed
\end{pf}

We now show $\prec$ is modular. Suppose \mbox{$x \in W_i$}, \mbox{$y
\in W_j$}, and \mbox{$x \prec y$}. Then \mbox{$i \leq j$} by
Lemma~\ref{lem-prec-imp-leq}. Suppose \mbox{$z \in W_k$}. Then
\mbox{$i \leq k$} or \mbox{$k \leq j$} by the modularity of
$\leq$. Suppose \mbox{$i < k$} or \mbox{$k < j$}. Then \mbox{$x \prec
z$} or \mbox{$z \prec y$} by condition~1. Otherwise \mbox{$i = k =
j$}, so \mbox{$x, y, z \in W_i$}. Since \mbox{$x \prec y$}, $W_i$ is
fully connected by condition~2, so \mbox{$x \prec z$} (and \mbox{$z
\prec y$}).

Finally, we show that $\prec$ is transitive. Suppose \mbox{$x \in
W_i$}, \mbox{$y \in W_j$}, \mbox{$z \in W_k$}, \mbox{$x \prec y$}, and
\mbox{$y \prec z$}. By Lemma~\ref{lem-prec-imp-leq}, \mbox{$i \leq j$}
and \mbox{$j \leq k$}, so \mbox{$i \leq k$} by the transitivity of
$\leq$. Suppose \mbox{$i < k$}. Then \mbox{$x \prec z$} by
condition~1. Otherwise \mbox{$i = k = j$}, so \mbox{$x, y, z \in
W_i$}. Since \mbox{$x \prec y$}, $W_i$ is fully connected by
condition~2, so \mbox{$x \prec z$}.
\qed
\end{pf}
(END OF PROPOSITION~\ref{prop-rep} PROOF)

\begin{prop}
	\mbox{$\mct \subset \mcb$} and is the set of reflexive
	relations in $\mcb$.
\end{prop}
\begin{pf}
We first show that \mbox{$\mct \subset \mcb$}. Let \mbox{$\preceq \in
\mct$} and \mbox{$x, y, z \in \mcw$}. By definition, $\preceq$ is
transitive. Suppose \mbox{$x \preceq y$}. Since $\preceq$ is total,
\mbox{$x \preceq z$} or \mbox{$z \preceq x$}. If \mbox{$z \preceq x$},
then \mbox{$z \preceq y$} by transitivity, so $\preceq$ is modular.
On the other hand, the empty relation over $\mcw$ is modular and
transitive, but not total and, consequently, not in $\mct$.

Now we show that \mbox{$\prec \in \mcb$} is in $\mct$ if and only if
it is reflexive. If \mbox{$\prec \in \mct$}, it is total, so it is
reflexive. If, instead, $\prec$ is reflexive, then \mbox{$x \prec x$}
so, by modularity, \mbox{$x \prec y$} or \mbox{$y \prec x$}. Thus,
$\prec$ is total. And, since \mbox{$\prec \in \mcb$}, it is
transitive.
\qed
\end{pf}

\begin{prop}\hspace{0ex}\\[-0.5cm]
	\begin{enumerate}

	\item \mbox{$\mcq \cap \mcb = \mct$}.

	\item \mbox{$\mcb \not\subseteq \mcq$}.

	\item \mbox{$\mcq \not\subseteq \mcb$} if $\mcw$ has at least
	three elements.

	\item \mbox{$\mcq \subset \mcb$} if $\mcw$ has one or two
	elements.

	\end{enumerate}
\end{prop}
\begin{pf}\hspace{0ex}\\[-0.5cm]
\begin{enumerate}

\item Suppose \mbox{$\preceq \in \mcq \cap \mcb$}. Then $\preceq$ is
total and transitive and, hence, in $\mct$. Suppose \mbox{$\preceq \in
\mct$}. By definition, $\preceq$ is total. Also by definition, it is
transitive, so by Proposition~\ref{prop-Props}, it is quasi-transitive
and, thus, in $\mcq$.  By Proposition~\ref{prop-TPsubMT},
\mbox{$\preceq \in \mcb$} and, so, in \mbox{$\mcq \cap \mcb$}.

\item The empty relation is modular and transitive, but not total and,
so, not in $\mcq$.

\item Suppose $a$ and $b$ are distinct elements of $\mcw$. The
relation \mbox{$\mcw \times \mcw \setminus \{(b,a)\}$} is total, and, since
the strict version is \mbox{$\{(a,b)\}$} which is transitive, it is also
quasi-transitive. However, if there are at least three elements in
$\mcw$, it is not transitive and, so, not in $\mcb$.

\item Suppose $\mcw$ has one element. Then $\mcb$ contains both
possible relations over $\mcw$, whereas $\mcq$ contains only the
fully connected relation over $\mcw$.

Suppose $\mcw$ has two elements $a$ and $b$. Then $\mcb$ contains
empty relation, the fully connected relation, and all the remaining
eight relations which contain either \mbox{$(a,b)$} or \mbox{$(b,a)$},
but not both.  $\mcq$, on the other hand, only contains the three
reflexive relations containing either \mbox{$(a,b)$} or
\mbox{$(b,a)$}.

\end{enumerate}
\qed
\end{pf}

\begin{prop}\hspace{0ex}\\[-0.5cm]
	\begin{enumerate}

	\item \mbox{$\mcsq \cap \mcb = \mcst$}.

	\item \mbox{$\mcb \not\subseteq \mcsq$}.

	\item \mbox{$\mcsq \not\subseteq \mcb$} if $\mcw$ has at least
	three elements.

	\item \mbox{$\mcsq \subset \mcb$} if $\mcw$ has one or two
	elements.

	\end{enumerate}
\end{prop}
\begin{pf}\hspace{0ex}\\[-0.5cm]
\begin{enumerate}

\item Suppose \mbox{$\prec \in \mcsq \cap \mcb$}. Since $\prec \in
\mcsq$, it is irreflexive, so since it is in $\mcb$, it is in $\mcst$
by Proposition~\ref{prop-TPisoMTI}.  Suppose, instead, \mbox{$\prec
\in \mcst$}. By Proposition~\ref{prop-TPisoMTI}, \mbox{$\prec \in
\mcb$}. Let \mbox{$\preceq \in \mct$} be a relation such that $\prec$
is its strict version. (Obviously such a relation must exist.)  From
Proposition~\ref{prop-MTandTQ}, \mbox{$\preceq \in \mcq$}, so
\mbox{$\prec \in \mcsq$}. Thus, \mbox{$\prec \in \mcsq \cap \mcb$}.

\item The fully connected relation over $\mcw$ is in $\mcb$, but not
asymmetric and, so, not in $\mcsq$.

\item Suppose $a$ and $b$ are distinct elements of $\mcw$. If $\mcw$
has at least three elements, the relation \mbox{$\{(a,b)\}$} is not
modular and, thus, not in $\mcb$, yet it is the strict version of the
relation \mbox{$\mcw \times \mcw \setminus \{(b,a)\}$} which is total
and quasi-transitive (since \mbox{$\{(a,b)\}$} is transitive).

\item Suppose $\mcw$ has one element. Then $\mcb$ contains both
possible relations over $\mcw$, whereas $\mcsq$ contains only the empty
relation over $\mcw$.

Suppose $\mcw$ has two elements $a$ and $b$. Then $\mcb$ contains
empty relation, the fully connected relation, and all eight of the
remaining relations which contain either \mbox{$(a,b)$} or
\mbox{$(b,a)$}, but not both. $\mcsq$, on the other hand, only
contains the three irreflexive relations.

\end{enumerate}
\qed
\end{pf}

\begin{prop}
	If \mbox{$S \subseteq \mcs$}, then \mbox{$Un(S)$} is modular
	but not necessarily transitive.
\end{prop}
\begin{pf}
Let \mbox{$\prec = Un(S)$}. Suppose \mbox{$x, y, z \in \mcw$} and
\mbox{$x \prec y$}. Then there is some \mbox{$s \in S$} such that
\mbox{$x <^s y$}. By assumption, $<^s$ is modular, so \mbox{$x <^s z$}
or \mbox{$z <^s y$}. By the definition of \mbox{$Un(S)$}, \mbox{$x
\prec z$} or \mbox{$z \prec y$}, so $\prec$ is modular.

Suppose \mbox{$a, b, c \in \mcw$} and \mbox{$S = \{s_1, s_2\}$} such
that \mbox{$<^{s_1} = \{(a,b), (a,c)\}$} and \mbox{$<^{s_2} = \{(b,a),
(c,a)\}$}. \mbox{$Un(S)$} is not transitive.
\qed
\end{pf}

\begin{prop}
	If \mbox{$S \subseteq \mcs$}, then \mbox{$AGRUn(S) \in \mcb$}.
\end{prop}
\begin{pf}
The transitive closure of any relation is transitive. Since
\mbox{$Un(S)$} is modular, the transitive closure of \mbox{$Un(S)$} is
also modular by Proposition~\ref{prop-Props}.
\qed
\end{pf}

\begin{prop}
	If \mbox{$S \subseteq \mcs$} and $\sqsupseteq_S$ is a total
	order, then \mbox{$AGRRf(S) \in \mcb$}.
\end{prop}
\begin{pf}
We first prove that \mbox{$AGRRf(S)$} is modular. Note that the proof
does not depend on $\sqsupseteq_S$ being a total order.
\begin{lem}\label{lem-AGRRfMod}
	If \mbox{$S \subseteq \mcs$}, then \mbox{$AGRRf(S)$} is
	modular.
\end{lem}
\begin{pf}
Let \mbox{$\prec = AGRRf(S)$}. Suppose \mbox{$x, y, z \in \mcw$} and
\mbox{$x \prec y$}. Then there exists \mbox{$s \in S$} such that
\mbox{$x <^s y$} and, for every \mbox{$s' \in S$}, \mbox{$s' \sqsupset
s$} implies \mbox{$x \not<^{s'} y$} and \mbox{$y \not<^{s'}
x$}. Consider a source $s'$ such that \mbox{$x <^{s'} z$}, \mbox{$z
<^{s'} x$}, \mbox{$y <^{s'} z$}, or \mbox{$z <^{s'} y$} and, for every
\mbox{$s'' \in S$}, \mbox{$s'' \sqsupset s'$} implies \mbox{$x
\not<^{s''} z$}, \mbox{$z \not<^{s''} x$}, \mbox{$y \not<^{s''} z$},
and \mbox{$z \not<^{s''} y$}. We know such a source exists since, by
modularity, \mbox{$x <^s z$} or \mbox{$z <^s y$}. Furthermore, since
$s'$ is a maximal rank such source, \mbox{$s' \sqsupseteq s$}. We
consider the four cases and show that, in each, either \mbox{$x \prec
z$} or \mbox{$z \prec y$}:

{\bf Case 1}: \mbox{$x <^{s'} z$}. Since, for every \mbox{$s'' \in
S$}, \mbox{$s'' \sqsupset s'$} implies \mbox{$x \not<^{s''} z$} and
\mbox{$z \not<^{s''} x$}, \mbox{$x \prec z$}.

{\bf Case 2}: \mbox{$z <^{s'} x$}. By modularity, \mbox{$z <^{s'} y$} or \mbox{$y <^{s'}
x$}.

Suppose \mbox{$z <^{s'} y$}. Since, for every \mbox{$s'' \in S$}, \mbox{$s'' \sqsupset s'$}
implies \mbox{$z \not<^{s''} y$} and \mbox{$y \not<^{s''} z$}, \mbox{$z \prec y$}.

Suppose \mbox{$y <^{s'} x$}. Then \mbox{$s \sqsupseteq s'$}, so
\mbox{$s \sqeq s'$}. Since \mbox{$x <^s y$}, \mbox{$x <^s z$} or
\mbox{$z <^s y$} by modularity. Thus, substituting $s$ for $s'$ above,
either \mbox{$x <^s z$} and, for every \mbox{$s'' \in S$}, \mbox{$s''
\sqsupset s$} implies \mbox{$x \not<^{s''} z$} and \mbox{$z
\not<^{s''} x$} so that \mbox{$x \prec z$}, or \mbox{$z <^s y$} and,
for every \mbox{$s'' \in S$}, \mbox{$s'' \sqsupset s$} implies
\mbox{$z \not<^{s''} y$} and \mbox{$y \not<^{s''} z$} so that \mbox{$z
\prec y$}.

{\bf Case 3}: \mbox{$y <^{s'} z$}. By modularity, \mbox{$y <^{s'} x$}
or \mbox{$x <^{s'} z$} which we have already considered in cases 2 and
1, respectively.

{\bf Case 4}: \mbox{$z <^{s'} y$}. We have already considered this in
case 2.
\qed
\end{pf}

It remains to show that $\prec$ is transitive. Suppose \mbox{$x \prec
y$} and \mbox{$y \prec z$}. Then there exists \mbox{$s_1 \in S$} such
that \mbox{$x <^{s_1} y$} and, for every \mbox{$s'_1 \in S$},
\mbox{$s'_1 \sqsupset s$} implies \mbox{$x \not<^{s'_1} y$} and
\mbox{$y \not<^{s'_1} x$}, and there exists \mbox{$s_2 \in S$} such
that \mbox{$y <^{s_2} z$} and, for every \mbox{$s'_2 \in S$, $s'_2
\sqsupset s$} implies \mbox{$y \not<^{s'_2} z$} and \mbox{$z
\not<^{s'_2} y$}. Suppose \mbox{$s_1 \sqsupset s_2$} (the case
\mbox{$s_2 \sqsupset s_1$} is similar). Then \mbox{$y \not<^{s_1} z$}
and \mbox{$z \not<^{s_1} y$}. By modularity, since \mbox{$x <^{s_1}
y$} and \mbox{$z \not<^{s_1} y$}, \mbox{$x <^{s_1} z$}. Let \mbox{$s'
\in S$} and \mbox{$s' \sqsupset s_1$}. Then \mbox{$x \not<^{s'} y$}
and \mbox{$y \not<^{s'} x$}. And, since \mbox{$s_1 \sqsupset s_2$},
\mbox{$s' \sqsupset s_2$}, so \mbox{$y \not<^{s'} z$} and \mbox{$z
\not<^{s'} y$}. By modularity, \mbox{$x \not<^{s'} z$} and \mbox{$z
\not<^{s'} x$}. Therefore, \mbox{$x \prec z$}.
\qed
\end{pf}

\begin{prop}
	If \mbox{$S \subseteq \mcs$}, then \mbox{$AGR^*(S) \in \mcb$}.
\end{prop}
\begin{pf}
By Proposition~\ref{prop-AGRUn}, \mbox{$\prec_r \in \mcb$} for every
\mbox{$r \in ranks(S)$}. For convenience, we assume the existence of a
``virtual'' source $s_r$ corresponding to each $\prec_r$. Precisely,
for each \mbox{$r \in ranks(S)$}, assume there exists a source
\mbox{$s_r \in \mcs$} such that \mbox{$<^{s_r} = \prec_r$} and
\mbox{$rank(s_r) = r$}, and let $S'$ be the set of these
sources. Then,
\begin{eqnarray*}
	\lefteqn{AGR^*(S)} \\
		& = & \left\{(x, y) :
			\begin{array}{l}
				\exists r \in \mcr.\ x <_r y \wedge \\
				\left(\forall r' > r \in ranks(S).\ x
					\sagnost_{r'} y\right)
			\end{array}
			\right\} \\
		& = & \left\{(x, y) :
			\begin{array}{l}
				\exists s \in S'.\ x <^s y \wedge \\
				\left(\forall s' \sqsupset s \in S'.\ x
					\sagnost^{s'} y\right)
			\end{array}
			\right\} \\
		& = & AGRRf(S').	
\end{eqnarray*}
Since there is one source in $S'$ per rank $r$, and since $>$ is a
total order over $\mcr$, $\sqsupseteq_{S'}$ is a total order. The
result follows from Proposition~\ref{prop-AGRRf}.
\qed
\end{pf}

\begin{prop}
	If \mbox{$S \subseteq \mcs$}, then \mbox{$AGR(S) \in \mcb$}.
\end{prop}
\begin{pf}
By Lemma~\ref{lem-AGRRfMod}, \mbox{$AGRRf(S)$} is modular.
\mbox{$AGRRf(S)^+$} is obviously transitive, and, by
Proposition~\ref{prop-Props}, it is modular as well.
\qed
\end{pf}

\begin{prop}
	Suppose \mbox{$S \subseteq \mcs$}.
	\begin{enumerate}

	\item If $\sqsupseteq_S$ is fully connected, \mbox{$AGR(S) =
	AGRUn(S)$}.

	\item If \mbox{$\sqsupseteq_S$} is a total order,
	\mbox{$AGR(S) = AGRRf(S)$}.

	\end{enumerate}
\end{prop}
\begin{pf}\hspace{0ex}\\[-0.5cm]
\begin{enumerate}

\item Suppose $\sqsupseteq_S$ is fully connected. Then the second half
of the definition of \mbox{$AGRRf$} is vacuously true so that
\mbox{$AGRRf(S)$} simplifies to \mbox{$\{(x,y) : \exists s \in S.\ x
<^s y\}$}. But this is exactly \mbox{$\bigcup_{s \in S} <^s$}, i.e.,
\mbox{$Un(S)$}, so \mbox{$AGR(S) = AGRRf(S)^+ = Un(S)^+ = AGRUn(S)$}.

\item Suppose $\sqsupseteq_S$ is an total order. By
Proposition~\ref{prop-AGRRf}, \mbox{$AGRRf(S)$} is transitive, so
\mbox{$AGR(S) = AGRRf(S)^+ = AGRRf(S)$}.

\end{enumerate}
\qed
\end{pf}

\begin{prop}
	Let \mbox{$S = \{s_1, \ldots, s_n\} \subseteq \mcs$} and
	\mbox{$AGR_f(<^{s_1}, \ldots, <^{s_n}) =
	AGR(S)$}. \mbox{$AGR_f$} satisfies (the modified versions of)
	restricted range, unrestricted domain, Pareto principle, IIA,
	and non-dictatorship.
\end{prop}
\begin{pf}
Let \mbox{$\prec = AGR_f(<^{s_1}, \ldots, <^{s_n})$}. Then
\mbox{$\prec = AGR(S)$}.

{\bf Restricted range:} \mbox{$AGR_f$} satisfies restricted range by
Proposition~\ref{prop-AGR}.

{\bf Unrestricted domain:} \mbox{$AGR_f$} satisfies unrestricted
domain by Definition~\ref{defn-sources}.

{\bf Pareto principle:} Suppose \mbox{$x <^{s_i} y$} for all $s_i$. In
particular, \mbox{$x <^s y$} where $s$ is a maximal rank source of
$S$. Since $s$ is maximal, it is vacuously true that for every
\mbox{$s' \sqsupset s \in S$}, \mbox{$x \not<^s y$} and \mbox{$y
\not<^{s'} x$}. Therefore, \mbox{$x \prec y$}, so \mbox{$AGR_f$}
satisfies the Pareto principle.

{\bf IIA:} Let \mbox{$S' = \{s'_1, \ldots, s'_n\}$}. First note that
\mbox{$AGRRf$} satisfies IIA:
\begin{lem}\label{lem-AGRRf-IIA}
	Suppose \mbox{$S = \{s_1, \ldots, s_n\} \subseteq \mcs$},
	\mbox{$S' = \{s'_1, \ldots, s'_n\} \subseteq \mcs$},
	\mbox{$s_i \sqeq s'_i$} for all $i$, \mbox{$\prec_* =
	AGRRf(S)$}, and \mbox{$\prec'_* = AGRRf(S')$}. If, for
	\mbox{$x, y \in \mcw$}, \mbox{$x <^{s_i} y$} iff \mbox{$x
	<^{s'_i} y$} for all $i$, then \mbox{$x \prec_* y$} iff
	\mbox{$x \prec'_* y$}.
\end{lem}
\begin{pf}
Suppose \mbox{$s_i \sqeq s'_i$}, and \mbox{$x <^{s_i} y$} iff \mbox{$x
<^{s'_i} y$}, for all $i$. Then \mbox{$x \prec_* y$} iff \mbox{$x
\prec'_* y$} since Definition~\ref{def-AGRRf} only relies on the
relative ranking of the sources and the relations between $x$ and $y$
in their belief states to determine the relation between $x$ and $y$
in the aggregated state.
\qed
\end{pf}
Thus, IIA can only be disobeyed when the closure step of \mbox{$AGR$}
introduces new opinions. (Note that IIA is satisfied when there are no
sources of equal rank since, by Proposition~\ref{prop-AGRSpecial}, the
closure step introduces no new opinions under these conditions.)
However, these new opinions are only added between worlds already
involved in a conflict, as the following two lemmas show:
\begin{lem}\label{lem-conflict1}
	Suppose \mbox{$S \subseteq \mcs$} and \mbox{$\prec_* =
	AGRRf(S)$}. For every integer \mbox{$n \geq 2$}, if \mbox{$x,
	y \in \mcw$}, \mbox{$x \not\prec_* y$}, there exist
	\mbox{$x_0, \ldots, x_n \in \mcw$} such that \mbox{$x = x_0
	\prec_* \cdots \prec_* x_n = y$}, and $n$ is the smallest
	integer such that this is true, then \mbox{$x_n \prec_* \cdots
	\prec_* x_0$}.
\end{lem}
\begin{pf}
Suppose \mbox{$x, y \in \mcw$}, \mbox{$x \not\prec_* y$}, and there
exist \mbox{$x_0, \ldots, x_n \in \mcw$} such that \mbox{$x = x_0
\prec_* \cdots \prec_* x_n = y$}, and $n$ is the smallest integer such
that this is true. Consider any triple \mbox{$x_{i-1}, x_i, x_{i+1}$},
where \mbox{$1 \leq i \leq n-1$}. First, \mbox{$x_{i-1} \not\prec_*
x_{i+1}$}, otherwise there would be a chain of shorter length than $n$
between $x$ and $y$. Now, since \mbox{$x_{i-1} \prec_* x_i$}, there
exists \mbox{$s_1 \in S$} such that \mbox{$x_{i-1} <^{s_1} x_i$} and,
for all \mbox{$s' \sqsupset s_1 \in S$}, \mbox{$x_{i-1} \sagnost^{s'}
x_i$}. Similarly, there exists \mbox{$s_2 \in S$} such that \mbox{$x_i
<^{s_2} x_{i+1}$} and, for all \mbox{$s' \sqsupset s_2 \in S$},
\mbox{$x_i \sagnost^{s'} x_{i+1}$}. Thus, all sources with higher rank
than \mbox{$\max(s_1, s_2)$} are agnostic with respect to $x_{i-1}$
and $x_{i+1}$.

Suppose \mbox{$s_1 \sqsupset s_2$}. Then \mbox{$x_i \sagnost^{s_1}
x_{i+1}$} so, by transitivity, \mbox{$x_{i-1} <^{s_1} x_{i+1}$}. But
then \mbox{$x_{i-1} \prec_* x_{i+1}$}, a contradiction. Similarly, we
derive a contradiction if \mbox{$s_2 \subset s_1$}. Thus, \mbox{$s_1
\sqeq s_2$}.

Now, since \mbox{$x_{i-1} \not\prec_* x_{i+1}$} and all sources with
rank higher than $s_1$ and $s_2$ are agnostic with respect to
$x_{i-1}$ and $x_{i+1}$, \mbox{$x_{i-1} \not<^{s_1} x_{i+1}$}. By
modularity, \mbox{$x_{i+1} <^{s_1} x_i$}. Since \mbox{$s_1 \sqeq
s_2$}, and all the sources with higher rank than $s_2$ are agnostic
with respect to $x_i$ and $x_{i+1}$, \mbox{$x_{i+1} \prec_*
x_i$}. Similarly, \mbox{$x_i <^{s_2} x_{i-1}$}, so \mbox{$x_i \prec_*
x_{i-1}$}.  Since $i$ was chosen arbitrarily between $1$ and
\mbox{$n-1$}, \mbox{$x_n \prec_* \cdots \prec_* x_0$}. And, in fact,
all the opinions between these worlds originate from sources of the
same rank.
\qed
\end{pf}

\begin{lem}\label{lem-conflict2}
	Suppose \mbox{$S \subseteq \mcs$}, \mbox{$\prec_* =
	AGRRf(S)$}, \mbox{$\prec = AGR(S)$}, and \mbox{$x \not\prec_*
	y$} for \mbox{$x, y \in \mcw$}. If \mbox{$x \prec y$}, then
	\mbox{$x \conflict y$}.
\end{lem}
\begin{pf}
Suppose \mbox{$x \not\prec_* y$}. If \mbox{$x \prec y$}, then there
exist \mbox{$x_0, \ldots, x_n$} such that \mbox{$x = x_0 \prec_*
\cdots \prec_* x_n = y$} and $n$ is the smallest positive integer such
that this is true. Then, by Lemma~\ref{lem-conflict1}, \mbox{$y = x_n
\prec_* \cdots \prec_* x_0 = x$}, so \mbox{$y \prec x$} and \mbox{$x
\conflict y$}.
\qed
\end{pf}

Now, suppose \mbox{$x, y \in \mcw$}, \mbox{$x <^{s_i} y$} iff \mbox{$x
<^{s_i'} y$} for all $i$, \mbox{$x \notconflict y$}, and \mbox{$x
\notconflict' y$}. We show that \mbox{$x \prec y$} implies \mbox{$x
\prec' y$} (the other direction is identical). Suppose \mbox{$x \prec
y$}. Let \mbox{$\prec_* = AGRRf(S)$} and \mbox{$\prec'_* =
AGRRf(S')$}. Since \mbox{$x \notconflict y$}, \mbox{$x \prec_* y$} by
Lemma~\ref{lem-conflict2}. But then \mbox{$x \prec'_* y$} by
Lemma~\ref{lem-AGRRf-IIA}, so \mbox{$x \prec' y$}. \\
(END OF IIA SUB-PROOF)

{\bf Non-dictatorship:} Suppose $\sqsupseteq_S$ is fully connected and
suppose \mbox{$x <^{s_i} y$} and \mbox{$y \not<^{s_i} x$}. Let $s_j$
be such that \mbox{$y <^{s_j} x$}. Then \mbox{$x \prec y$} and
\mbox{$y \prec x$}, so $s_i$ is not a dictator.
\qed
\end{pf}
(END OF PROPOSITION~\ref{prop-Arrow2} PROOF)

\begin{prop}
	Let $\mca$ and $S$ be as in Definition~\ref{def-fusion},
	$\prec^{A_i}$, agent $A_i$'s induced belief state, and
	$\sqsupseteq_S$, fully connected. If \mbox{$A = \fuse(\mca)$},
	then \mbox{$\left(\bigcup_{A_i \in \mca}
	\prec^{A_i}\right)^+$} is $A$'s induced belief state.
\end{prop}
\begin{pf}
We will use the following lemma:
\begin{lem}\label{prop-RedundantTC}
	If $\Pi$ is a set of relations over an arbitrary finite set
	$\Omega$, then
	\begin{displaymath}
		\left(\bigcup_{\leq \in \Pi} \leq^+\right)^+ =
		\left(\bigcup_{\leq \in \Pi} \leq\right)^+
	\end{displaymath}
	where $\leq^+$ is the transitive closure of $\leq$.
\end{lem}
\begin{pf}
Let \mbox{$\prec = \left(\bigcup_{\leq \in \Pi} \leq^+\right)^+$},
\mbox{$\prec' = \left(\bigcup_{\leq \in \Pi} \leq\right)^+$}, and
\mbox{$a, b \in \Omega$}.  Suppose \mbox{$a \prec b$}. Then there
exist \mbox{$\leq_0, \ldots, \leq_{n-1} \in \Pi$} and \mbox{$w_0,
\ldots, w_n \in \Omega$} such that
\begin{displaymath}
	a = w_0 \leq_0^+ \cdots \leq_{n-1}^+ w_n = b
\end{displaymath}
Thus, there exist \mbox{$x_{00}, \ldots, x_{0m_0}, \ldots$},
\mbox{$x_{(n-1)0}, \ldots, x_{(n-1)m_{n-1}}$} in $\Omega$ such that
\begin{eqnarray*}
	a & = & w_0 = x_{00} \leq_0 \cdots \leq_0 x_{0m_0} = w_1
		= x_{10} \cdots\\
	  & = & w_{n-1} = x_{(n-1)0} \leq_{n-1} \cdots \leq_{n-1}
		x_{(n-1)m_{n-1}}\\
	  & = & w_n = b 
\end{eqnarray*}
so \mbox{$a \leq' b$}.

Now suppose \mbox{$a \prec' b$}. Then there exist \mbox{$\leq_0,
\ldots, \leq_{n-1} \in \Pi$} and \mbox{$w_0, \ldots, w_n \in \Omega$}
such that
\begin{displaymath}
	a = w_0 \leq_0 \cdots \leq_{n-1} w_n = b
\end{displaymath}
Obviously, this implies that
\begin{displaymath}
	a = w_0 \leq_0^+ \cdots \leq_{n-1}^+ w_n = b
\end{displaymath}
which implies that
\begin{displaymath}
	a = w_0 \leq_*^+ \cdots \leq_*^+ w_n = b
\end{displaymath}
where \mbox{$\leq_* = \left(\bigcup_{\leq \in \Pi} \leq\right)$}, so
\mbox{$a \leq b$}.
\qed
\end{pf}

Now, let $\prec$ be the belief state induced by
\mbox{$\fuse(\mca)$}. Then \mbox{$\prec = AGR(S)$}. By
Proposition~\ref{prop-AGRSpecial}, \mbox{$\prec = AGRUn(S)$}, so
\begin{eqnarray*}
	\prec	& = & Un(S)^+ \\
		& = & \left(\bigcup_{s \in S} <^s\right)^+ \\
		& = & \left(\bigcup_{s \in \bigcup^n_{i=1} S_i}
			<^s\right)^+ \\
		& = &\left(\bigcup_{A_i \in \mca}\bigcup_{s \in S_i}
			<^s\right)^+
\end{eqnarray*}
By the lemma,
\begin{eqnarray*}
	\prec	& = & \left(\bigcup_{A_i \in \mca}\left(\bigcup_{s \in
			S_i} <^s\right)^+\right)^+ \\
		& = & \left(\bigcup_{A_i \in \mca} AGRUn(S_i)\right)^+
			\\
		& = & \left(\bigcup_{A_i \in \mca} \prec^{A_i}\right)^+
\end{eqnarray*}
\qed
\end{pf}

\begin{prop}
	Let $A$ be an agent informed by a set of sources \mbox{$S
	\subseteq \mcs$} and with pedigreed belief state
	\mbox{$(\prec, l)$}. Then
	\begin{center}
	\begin{displaymath}
		x \prec^A_r y
	\end{displaymath}
	iff
	\begin{displaymath}
		\begin{array}{l}
		\exists s \in S.\ x <^s y \wedge r = rank(s) \wedge \\
		\left(\forall s' \sqsupset s \in S.\ x \sagnost^{s'}
			y\right).
		\end{array}
	\end{displaymath}
	\end{center}
\end{prop}
\begin{pf}
Suppose \mbox{$x \prec^A_r y$}. Then \mbox{$x \prec y$} and
\mbox{$l((x,y)) = r$}. By Definitions~\ref{def-AGRRf} and
\ref{def-PBS}, there exists \mbox{$s \in S$} such that \mbox{$x <^s
y$} and for every \mbox{$s' \sqsupset s \in S$}, \mbox{$x
\sagnost^{s'} y$}. In particular, if \mbox{$x <^{s'} y$} for some
\mbox{$s' \in S$}, then \mbox{$s \sqsupset s'$}, so \mbox{$rank(s)
\geq rank(s')$}. Thus,
\begin{eqnarray*}
	r & = & l((x,y)) \\
		& = & \max\{rank(s') : x <^{s'} y, s' \in S\} \\
		& = & rank(s).
\end{eqnarray*}

Now suppose there exists \mbox{$s \in S$} such that \mbox{$x <^s y$},
\mbox{$r = rank(s)$}, and, for every \mbox{$s' \sqsupset s \in S$},
\mbox{$x \sagnost^{s'} y$}. Then \mbox{$x \prec y$}. Moreover, since
for every \mbox{$s' \in S$}, \mbox{$x <^{s'} y$} implies \mbox{$s
\sqsupseteq s'$} which implies \mbox{$rank(s) \geq rank(s')$},
\begin{eqnarray*}
	l((x,y)) & = & \max\{rank(s') : x <^{s'} y, s' \in S\} \\
		& = & rank(s) \\
		& = & r.
\end{eqnarray*}
Therefore, \mbox{$x \prec^A_r y$}.
\qed
\end{pf}

\begin{prop}
	Let $\mca$ and $S$ be as in Definition~\ref{def-fusion},
	$\sqsupseteq_S$, a total pre-order, and \mbox{$A =
	\fuse(\mca)$}. If
	\begin{enumerate}

	\item $\prec$ is the relation
	\begin{displaymath}
		\left\{(x,y) :
		\begin{array}{l}
			\exists A_i \in \mca, r \in \mcr.\ x
				\prec^{A_i}_r y \wedge \\
			\left(\forall A_j \in \mca, r' > r \in
				\mcr.\ x \agnost^{A_j}_{r'} y\right)
		\end{array}
		\right\} 
	\end{displaymath}
	over $\mcw$,

	\item \mbox{$l : \prec \to \mcr$} such that \mbox{$l((x,y)) =
	\max\{r : x \prec^{A_i}_r y, A_i \in \mca\}$}, and

	\end{enumerate}
	then \mbox{$(\prec, l)$} is $A$'s pedigreed belief state.
\end{prop}
\begin{pf}
Let \mbox{$\prec' = AGRRf(S)$} and \mbox{$l' : \prec' \to \mcr$} such
that \mbox{$l'((x,y)) = \max\{rank(s) : x <^s y, s \in S\}$}. It
suffices to show that \mbox{$\prec = \prec'$} and \mbox{$l = l'$}.

Suppose \mbox{$x \prec y$}. We show that \mbox{$x \prec' y$}, i.e.,
there exists \mbox{$s \in S$} such that \mbox{$x <^s y$} and, for
every \mbox{$s' \sqsupset s \in S$}, \mbox{$x \not<^{s'} y$} and
\mbox{$y \not<^{s'} x$}, and that \mbox{$l'((x,y)) = l((x,y))$}. Since
\mbox{$x \prec y$}, there exists $A_i$ and $r$ such that \mbox{$x
\prec^{A_i}_r y$} and, for every \mbox{$A_j \in \mca$} and \mbox{$r' >
r \in \mcr$}, \mbox{$x \not\prec^{A_j}_{r'} y$} and \mbox{$y
\not\prec^{A_j}_{r'} x$}. Since \mbox{$x \prec^{A_i}_r y$}, there
exists \mbox{$s \in S_i$} such that \mbox{$x <^s y$}, \mbox{$rank(s) =
r$}, and, for every \mbox{$s_1 \sqsupset s \in S_i$}, \mbox{$x
\not<^{s_1} y$} and \mbox{$y \not<^{s_1} x$}. \mbox{$S_i \subseteq
S$}, so there exists \mbox{$s \in S$} such that \mbox{$x <^s y$}. Now
suppose $s'$ is a maximal rank source of $S$ with \mbox{$x <^{s'} y$}
or \mbox{$y <^{s'} x$}. Such an $s'$ exists since \mbox{$x <^s
y$}. Since $\sqsupseteq$ is a total pre-order, it suffices to show that
\mbox{$s \sqsupseteq s'$}. Suppose \mbox{$s' \in S_j$}. Since
\mbox{$S_j \subseteq S$}, $s'$ is also a maximal rank source of $S_j$
with \mbox{$x <^{s'} y$} or \mbox{$y <^{s'} x$}, so \mbox{$x
\prec^{A_j}_{rank(s')} y$} or \mbox{$y \prec^{A_j}_{rank(s')} x$}. But
since \mbox{$x \prec^{A_i}_r y$}, \mbox{$r = rank(s) \geq rank(s')$},
so \mbox{$s \sqsupseteq s'$}. Furthermore, \mbox{$l'((x,y)) = rank(s)
= r = l((x,y))$}.

Now suppose \mbox{$x \prec' y$}. We show that \mbox{$x \prec y$},
i.e., there exists $A_i$ and $r$ such that \mbox{$x \prec^{A_i}_r y$}
and, for every \mbox{$A_j \in \mca$} and \mbox{$r' > r \in \mcr$},
\mbox{$x \not\prec^{A_j}_{r'} y$} and \mbox{$y \not\prec^{A_j}_{r'}
x$}, and that \mbox{$l((x,y)) = l'((x,y))$}. Since \mbox{$x \prec'
y$}, there exists \mbox{$s \in S$} such that \mbox{$x <^s y$} and, for
every \mbox{$s' \sqsupset s \in S$}, \mbox{$x \not<^{s'} y$} and
\mbox{$y \not<^{s'} x$}. Suppose \mbox{$s \in S_i$}. Since \mbox{$S_i
\subseteq S$}, it is also the case that for every \mbox{$s' \sqsupset
s \in S_i$}, \mbox{$x \not<^{s'} y$} and \mbox{$y \not<^{s'} x$}, so
\mbox{$x \prec^{A_i}_{rank(s)} y$}. Now, let $A_j$ and $r'$ be such
that \mbox{$x \prec^{A_j}_{r'} y$} or \mbox{$y \prec^{A_j}_{r'}
x$}. It suffices to show that \mbox{$rank(s) \geq r'$}. By
Proposition~\ref{prop-label}, there exists \mbox{$s' \in S_j$} such
that \mbox{$x <^{s'} y$} or \mbox{$y <^{s'} x$} and \mbox{$rank(s') =
r'$}. But then \mbox{$s \sqsupseteq s'$}, so \mbox{$rank(s) \geq
rank(s') = r'$}. Furthermore, \mbox{$l((x,y)) = rank(s) = l'((x,y))$}.
\qed
\end{pf}

\begin{prop}
	If $\mca$ and $S$ are as in Definition~\ref{def-fusion},
	$\sqsupseteq_S$ is a total order, and \mbox{$(\prec, l)$} is
	the pedigreed belief state of \mbox{$\fuse(\mca)$}, then
	\mbox{$\prec^+ = \prec$}.
\end{prop}
\begin{pf}
Since $\sqsupseteq_S$ is a total order, \mbox{$AGR(S) = AGRRf(S)$} by
Proposition~\ref{prop-AGRSpecial}. Thus, \mbox{$\prec = AGRRf(S) =
AGR(S) = AGRRf(S)^+ = \prec^+$}.
\qed
\end{pf}

\newpage

\section{Notation key}
%	 ____________

$\Omega$: arbitrary finite set \\
$a, b, c, \ldots$: specific elements of a set \\
$x, y, z, \ldots$: arbitrary elements of a set \\
$A, B, C, \ldots$: specific subsets of a set \\
$X, Y, Z, \ldots$: arbitrary subsets of a set \\
$\Pi$: arbitrary set of relations \\
$\leq$: arbitrary relation \\
$\C(X, \leq)$: choice set of $W$ wrt $\leq$ \\

\noindent
$\mcw$: finite set of possible worlds/alternatives \\
$w$, $W$: element, subset of $\mcw$, respectively \\

\noindent
$\mcb$: set of generalized belief states (modular, transitive
	relations) \\
$\prec$: element of $\mcb$, strict likelihood/preference \\
$\preceq$: weak likelihood/preference \\
$\agnost$: equal likelihood, agnosticism/indifference \\
$\conflict$: conflict \\

\noindent
$\mct$: set of total pre-orders \\
$\mcst$: strict versions of total pre-orders \\
$\mcq$: set of total, quasi-transitive relations \\
$\mcsq$: strict versions of total, quasi-transitive relations \\

\noindent
$\mcs$: set of sources \\
$s$, $S$: element, subset of $\mcs$, respectively \\
$<^s$: belief state of source $s$ \\
$\sagnost$: source agnosticism \\
$\sconflict$: source conflict \\
$\mcr$: set of ranks \\
$r$: element of $\mcr$ \\
$\sqsupseteq$, $\sqsupseteq_S$: credibility ordering over $\mcs$, $S
	\subseteq \mcs$, respectively \\

\noindent
$\mca$: set of agents \\
$A$: element of $\mca$ \\
$\prec^A$: $A$'s induced belief state \\
$(\prec, l)$: pedigreed belief state \\
$l$: labeling function of a pedigreed belief state \\
$\prec^A_r$: restriction of $A$'s pedigreed belief state to rank $r$ \\

\end{appendix}

\end{document}